\documentclass[11pt,a4paper]{article}
\usepackage[margin=25mm]{geometry}
\usepackage{times}
\usepackage[authoryear,round]{natbib}
\usepackage{amsmath,amsfonts,bm}

\def\eqref#1{equation~\ref{#1}}

\def\1{\bm{1}}

\DeclareMathAlphabet{\mathsfit}{\encodingdefault}{\sfdefault}{m}{sl}
\SetMathAlphabet{\mathsfit}{bold}{\encodingdefault}{\sfdefault}{bx}{n}

\newcommand{\E}{\mathbb{E}}

\newcommand{\Var}{\mathrm{Var}}

\usepackage[utf8]{inputenc}
\usepackage[T1]{fontenc}
\usepackage{hyperref}

\usepackage{url}
\usepackage{amsmath,amssymb,amsthm}
\usepackage{booktabs}
\usepackage{multirow}
\usepackage{graphicx}
\usepackage[table]{xcolor}
\usepackage{tikz}
\usetikzlibrary{decorations.pathreplacing,arrows.meta}
\usepackage{microtype}
\usepackage{enumitem}
\usepackage{caption}
\usepackage{float}
\usepackage{placeins} 
\usepackage[capitalize,nameinlink]{cleveref}

\definecolor{wordc}{HTML}{EB6834}
\definecolor{freec}{HTML}{2A78D6}
\definecolor{pixc}{HTML}{52514E}
\definecolor{ascc}{HTML}{A86F00}
\definecolor{descc}{HTML}{13855C}
\definecolor{rungrow}{HTML}{EEF4FC}
\definecolor{baserow}{HTML}{F1F0EC}
\hypersetup{colorlinks=true,linkcolor=freec!70!black,citecolor=freec!70!black,urlcolor=freec!70!black,
pdftitle={Making Language the Low-Rate Code of Vision: Successive Refinement from Words to Pixels},
pdfauthor={Ruilin Gong and Hao Gong}}

\newcommand{\rung}{\textsc{Rung}}

\newcommand{\cT}{\mathcal{T}}
\newcommand{\hR}{\hat R}
\newcommand{\hH}{\hat H}
\newcommand{\ind}{\mathbf{1}}
\newcommand{\Lrec}{\mathcal{L}_{\mathrm{rec}}}
\newcommand{\Llang}{\mathcal{L}_{\mathrm{lang}}}
\newcommand{\Lcons}{\mathcal{L}_{\mathrm{cons}}}

\newtheorem{proposition}{Proposition}
\newtheorem{corollary}{Corollary}
\theoremstyle{definition}

\crefname{proposition}{Proposition}{Propositions}
\crefname{corollary}{Corollary}{Corollaries}

\setlist{leftmargin=1.2em,itemsep=1pt,topsep=2pt}

\title{Making Language the Low-Rate Code of Vision:\\ Successive Refinement from Words to Pixels}

\author{\begin{tabular}{c@{\hspace{1cm}}c}
Ruilin Gong & Hao Gong\\[0.4em]
\small Henan University of Technology & \small University of Chinese Academy of Sciences\\[0.2em]
\footnotesize\href{mailto:251170400101@stu.haut.edu.cn}{251170400101@stu.haut.edu.cn} &
\footnotesize\href{mailto:gonghao23@mails.ucas.ac.cn}{gonghao23@mails.ucas.ac.cn}
\end{tabular}}
\date{}

\begin{document}
\maketitle

\begin{abstract}
An ordered visual code lets a short prefix stand in for an image, and the training loss decides what that prefix contains. Under pixel reconstruction the first bits go to the directions of highest variance, and we prove that a caption can then be independent of the entire first half of a code that is rate-optimal for pixels at every length. In the rate--distortion sense a caption is therefore not the low-rate code of its image, unless the code is built to make it one. We study that construction and what it costs. We cast captioning and text-to-image generation as successive refinement under a distortion measure that changes along the code, from caption log-loss in the early stages to pixel distortion in the last, so that each stage is compared with its own optimum. Putting words first then costs nothing among the caption stages exactly when the captions form a summarization chain, and it costs the pixels at most the caption uncertainty left in a pixel-optimal reconstruction, which is large for short prefixes and small at full length. We build \rung, an ordered code whose prefixes are supervised by a summarization ladder of captions and whose full length reconstructs the image; generation extends the code past the prompt and captioning truncates it. With data and capacity held fixed, ordering by language instead of reconstruction raises prefix--caption information $2.9\times$ and GenEval from 0.76 to 0.83, at a measured cost of 2.8\,dB PSNR at a quarter of the code and 0.1\,dB at full length. \rung-XL (3.4B) reaches 0.88 GenEval and 87.6 DPG-Bench in 32 network evaluations and 0.91 GenEval when errors are redrawn at the rung where they appear; its coarse and fine captions agree 94.1\% of the time, and it edits images without inversion.
\end{abstract}

\section{Introduction}\label{sec:intro}

Ordered visual codes make coarse-to-fine a property of the representation rather than of the sampler. Nested dropout \citep{rippel2014nested} and its recent descendants \citep{bachmann2025flextok,wen2025semanticist,miwa2025onedpiece} train one-dimensional tokenizers so that every prefix decodes to a valid image and a short prefix can stand in for the whole. Such a code has to decide what a reader who stops early should learn, and a reconstruction loss makes that decision by default: the first bits go to whatever reduces pixel error most, which in photographs means illumination and low-frequency colour.

The default matters most where text and images meet. Unified models for generation and understanding share one token stream \citep{wang2024emu3,chen2025janus}, diffuse over it \citep{yang2025mmada,xin2025dimoo}, pair a language model with a diffusion decoder \citep{zhou2025transfusion,deng2025bagel}, or transport text directly to pixels \citep{liu2025crossflow,he2025flowtok}. They differ in mechanism but share a premise, that text and images are peer modalities to be mapped onto each other. The premise overlooks a large difference in rate. A thirty-word caption carries about a hundred nats under a language model (\cref{app:laddercap}), a one-dimensional tokenizer spends a few thousand nats on a $256{\times}256$ image (2.5k in ours), and a codec aiming at faithful reconstruction spends several times more. Every text-to-image model therefore has to supply the missing information, and the models above supply it without structure, as Gaussian noise, sampling temperature or unmasking order. Nothing records which parts of the output the prompt fixed and which it left open. On GenEval \citep{ghosh2023geneval}, the baselines in \cref{tab:geneval} that score 0.80 or more overall reach 0.99--1.00 on single objects, yet all but one fall below 0.65 on counting or on position. Short prompts state these attributes explicitly, so the errors fall on exactly what the prompt specifies.

A natural fix is to put the caption at the front of an ordered code, so that generation extends the code and captioning truncates it. This is successive refinement \citep{equitz1991,rimoldi1994} across modalities, summed up by the slogan that language is the low-rate code of vision. Under the distortion measure that ordered codes are trained with, the slogan is false. \Cref{prop:pixel-first} constructs a source for which a code that is rate-optimal for pixels at every length carries \emph{no} caption information in its first half: squared error spends early rate on high-variance directions, and captions seldom describe them. On real images, a reconstruction-ordered code keeps 0.68 nats of caption information per caption token at a quarter of its length, against 1.97 when the order is anchored in language (\cref{fig:price}a).

We start from this failure. Which information goes first is a choice of distortion measure, made stage by stage, and making language the low-rate code of vision is one such choice: the early stages are judged by how well they determine captions and the last by how well it reconstructs pixels. Successive refinement under a distortion measure that changes along the code tells us when this choice is free and what it costs otherwise. Among the caption stages, words-first loses nothing exactly when the captions form a summarization chain (\cref{prop:ladder}). On the pixel side it costs an excess rate bounded by the caption uncertainty that a pixel-optimal reconstruction leaves open (\cref{prop:price}); this cost is paid at short prefixes, where the words are, and shrinks as fidelity grows. Taken together, words-first is free exactly when each caption summarizes the next finer one and some pixel-optimal reconstruction already determines the captions.

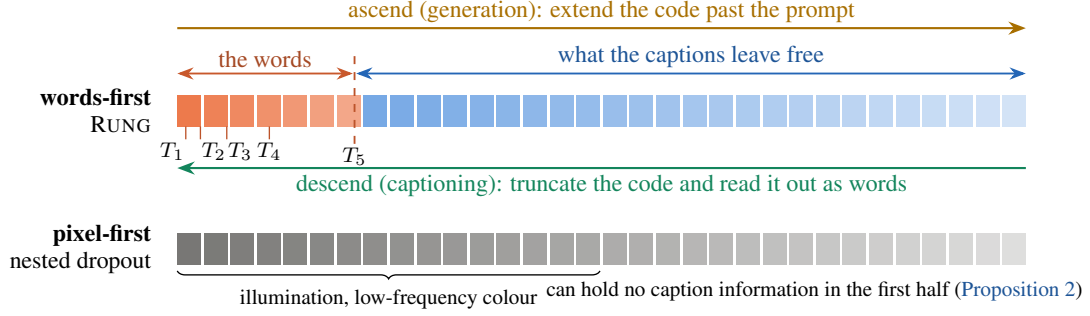
\begin{figure}[t]
\centering
\begin{tikzpicture}[x=0.352cm,y=1cm,font=\footnotesize,>=Stealth]
  \foreach \i in {0,...,31}{
    \ifnum\i<7
      \pgfmathsetmacro{\s}{88-5*\i}
      \fill[wordc!\s!white] (\i+0.06,2.25) rectangle (\i+0.94,2.67);
    \else
      \pgfmathsetmacro{\s}{64-1.8*(\i-7)}
      \fill[freec!\s!white] (\i+0.06,2.25) rectangle (\i+0.94,2.67);
    \fi}
  \foreach \t in {0.38,0.93,1.92,3.52}{\draw[wordc!80!black,line width=0.5pt] (\t,2.25) -- (\t,2.08);}
  \draw[wordc!80!black,dashed,line width=0.7pt] (6.72,2.02) -- (6.72,3.2);
  \node[anchor=north east,inner sep=1pt,font=\scriptsize] at (0.45,2.08) {$T_1$};
  \node[anchor=north west,inner sep=1pt,font=\scriptsize] at (0.86,2.08) {$T_2$};
  \node[anchor=north west,inner sep=1pt,font=\scriptsize] at (1.85,2.08) {$T_3$};
  \node[anchor=north,inner sep=1pt,font=\scriptsize] at (3.52,2.08) {$T_4$};
  \node[anchor=north,inner sep=1pt,font=\scriptsize] at (6.72,2.02) {$T_5$};
  \draw[<->,wordc!85!black,line width=0.7pt] (0.06,2.95) -- (6.66,2.95) node[midway,above,inner sep=2pt,text=wordc!70!black] {the words};
  \draw[<->,freec!85!black,line width=0.7pt] (6.78,2.95) -- (31.94,2.95) node[midway,above,inner sep=2pt,text=freec!70!black] {what the captions leave free};
  \draw[->,ascc,line width=0.8pt] (0.06,3.55) -- (31.94,3.55) node[midway,above,inner sep=2pt,text=ascc] {ascend (generation): extend the code past the prompt};
  \draw[<-,descc,line width=0.8pt] (0.06,1.7) -- (31.94,1.7) node[midway,below,inner sep=2pt,text=descc] {descend (captioning): truncate the code and read it out as words};
  \node[anchor=east,align=right,inner sep=0pt] at (-0.9,2.46) {\textbf{words-first}\\[-1pt] \rung};
  \foreach \i in {0,...,31}{
    \pgfmathsetmacro{\s}{78-1.9*\i}
    \fill[pixc!\s!white] (\i+0.06,0.42) rectangle (\i+0.94,0.84);}
  \draw[decorate,decoration={brace,mirror,amplitude=3pt},line width=0.5pt] (0.06,0.34) -- (15.94,0.34)
    node[midway,below=3pt,font=\scriptsize] {illumination, low-frequency colour};
  \node[font=\scriptsize,anchor=north] at (24,0.34) {can hold no caption information in the first half (\cref{prop:pixel-first})};
  \node[anchor=east,align=right,inner sep=0pt] at (-0.9,0.63) {\textbf{pixel-first}\\[-1pt] nested dropout};
\end{tikzpicture}
\caption{\textbf{Allocation of capacity in ordered visual codes.} Comparison between words-first (\textsc{Rung}, top) and standard pixel-first (nested dropout, bottom) ordering at identical code lengths and bitrates. In \textsc{Rung}, early rungs are supervised by natural language summaries whose boundaries $T_1,\dots,T_5$ are fixed by empirical rate matching, leaving subsequent rungs to encode visual variation unconstrained by text. Conversely, pixel-reconstruction objectives allocate early bits to high-variance global illumination and color, starving early rungs of caption information.}
\label{fig:overview}
\end{figure}

In return, the order provides three things that a map between peer modalities does not. The missing information has a place, the suffix, so the model knows what the prompt fixed and what it left free. Generation and captioning operate on one object, so caption--image agreement is a property of the code rather than of a cycle loss. And an error can be repaired at the rung that introduced it, before later rungs build on it. \Cref{fig:overview} shows the resulting model, \rung, next to the order it replaces. Our contributions are the following:
\begin{itemize}
\item \textbf{A distortion schedule for cross-modal coding.} We treat cross-modal generation as successive refinement under a schedule of distortion measures, which fixes the optimum each stage is compared with, and we give the condition under which a caption ladder costs nothing (\cref{prop:ladder}). The condition is a rule for building training data; violating it costs 5 GenEval points in our ablation.
\item \textbf{Both prices, bounded and measured.} A reconstruction order can leave language with nothing (\cref{prop:pixel-first}). A language order costs the pixels at most the caption uncertainty left in a pixel-optimal reconstruction, and this cost vanishes with fidelity (\cref{prop:price}). We measure 2.8\,dB at a quarter of the code and 2.9\% of the rate at full length, with realism unchanged (\cref{fig:price}).
\item \textbf{\rung.} A language-anchored ordered code with rate-matched cuts, deterministic descent, stochastic ascent and rung-level self-correction. With data, capacity and generator matched, the order alone is worth 7 GenEval points. \rung-XL reaches 0.88 GenEval in 32 network evaluations, level with the reward-tuned Lumina-DiMOO at 64, and 0.91 with rung-level correction at $1.2\times$ compute, where best-of-$N$ with the same verifier reaches 0.89 at $4\times$. The same code captions at a chosen granularity, edits without inversion, and transfers to radiology reports.
\end{itemize}

\section{Related work}\label{sec:related}

\textbf{Unified and coarse-to-fine generation.} Autoregressive \citep{wang2024emu3,chen2025janus,xie2025showo,xie2025showo2}, discrete-diffusion \citep{yang2025mmada,xin2025dimoo,shi2025muddit,li2025lavidao} and hybrid \citep{zhou2025transfusion,deng2025bagel,wu2025omnigen2} unified models place text and image tokens in one sequence and let the model infer how they relate \citep{zhao2025survey}. We place them at two depths of one code, so that the relation is part of the representation. Data-to-data couplings transport text directly to images \citep{liu2025crossflow,he2025flowtok} and close the entropy gap by injecting noise on the text side \citep{albergo2023}. We keep the gap and give it a location: it is the part of the code that the caption does not reach. Next-scale \citep{tian2024var}, bitwise \citep{han2025infinity}, cascaded \citep{ramesh2022dalle2} and masked \citep{chang2023muse} generators work coarse-to-fine over a spatial or spectral hierarchy on which text acts uniformly; ours is a rate hierarchy in which text occupies a specific interval.

\textbf{Ordered and language-aware tokenizers.} Nested dropout \citep{rippel2014nested} and Matryoshka embeddings \citep{kusupati2022matryoshka} make prefixes usable on their own, and FlexTok \citep{bachmann2025flextok}, Semanticist \citep{wen2025semanticist} and One-D-Piece \citep{miwa2025onedpiece} bring the idea to 1D image tokenizers \citep{yu2024titok}. They are our closest antecedents, and in all of them the order is set by reconstruction; Semanticist makes this explicit with a PCA-like ordering by explained variance. A second line makes codes semantic. SPAE and LQAE express images in a language model's vocabulary \citep{yu2023spae,liu2023lqae}; VILA-U, UniTok, TokenFlow and QLIP align codes with text-aligned features \citep{wu2025vilau,ma2025unitok,qu2025tokenflow,zhao2025qlip}; TA-TiTok conditions detokenization on text \citep{kim2025tatitok}. These methods change what tokens mean but leave the order to reconstruction, and none of them ties a prefix to a caption of matched rate (cf.\ the DINOv3 row of \cref{tab:order}).

\textbf{Rate--distortion with semantic targets.} Successive refinement \citep{equitz1991,rimoldi1994} asks when one nested code is optimal at every rate for a distortion measure. Remote source coding measures distortion on a hidden variable observed through the source \citep{dobrushin1962,witsenhausen1980} and becomes the information bottleneck under log-loss \citep{tishby1999ib,harremoes2007ib,courtade2014logloss}. Semantic rate--distortion constrains semantic and appearance distortion at a single rate \citep{liu2021semanticrd}, and rate--distortion--perception theory adds realism \citep{blau2019rdp,zhang2021universal}. Closest to our setting is scalable coding for humans and machines \citep{choi2022scalable,harell2025rdmachines}, whose base layer serves a machine task. We let the target change \emph{along} one nested code and quantify what each order costs; our base layer is language, which is also the generator's input.

\section{Ordering a code by a distortion schedule}\label{sec:theory}

\subsection{Ordered codes and distortion schedules}\label{sec:schedule}

Let $X$ be an image (its VAE latent) and $T_1,\dots,T_M$ captions from coarse to fine. We take captions to be deterministic functions of the image, $T_m=\psi_m(X)$, which holds when a fixed captioner and summarizer decode greedily (\cref{sec:setup}); \cref{app:stochastic} treats sampled captions. An ordered code maps $X$ to $Z=(Z_1,\dots,Z_L)$, and a reader who stops at $k$ sees the prefix $Z_{\le k}$. Rates are informational, $R_k=I(X;Z_{\le k})$. The successive-refinement coding theorem makes them operational for block codes \citep{rimoldi1994,elgamal2011nit}, and \cref{sec:cuts} measures them for learned codes.

An order is determined by what each stage is judged on. A \emph{distortion schedule} makes this explicit by attaching a distortion measure to every stage and comparing each stage with its own optimum:
\begin{itemize}
\item a \emph{caption stage} $m$ is judged by log-loss on $T_m$. With deterministic captions a prefix is lossless for $T_m$ when it determines $T_m$, and the least rate at which any code does so is $H(T_m)$;
\item a \emph{pixel stage} is judged by a distortion $d(X,\hat X)$ and compared with the rate--distortion function $R_X(D)$, the least $I(X;\hat X)$ subject to $\E d(X,\hat X)\le D$. The theory uses squared error; \cref{sec:price} also reports perceptual metrics.
\end{itemize}
A nested code is \emph{successively refinable for a schedule} if every stage attains its own optimum simultaneously, and the \emph{excess rate} of a stage is its rate beyond that optimum. Classical successive refinement uses one measure throughout. With a pixel measure this is the order of existing ordered tokenizers, which we call \emph{pixel-first}. \emph{Words-first} has caption stages $1,\dots,M$ followed by pixel stages. Neither order is intrinsic to the image, and \cref{sec:two-prices} shows that each has a cost.

\subsection{When the caption stages cost nothing}\label{sec:caption-stages}

\begin{proposition}[Caption ladder]\label{prop:ladder}
Let a nested code determine $T_m$ from its stage-$m$ prefix for every $m$. Its stage-$m$ rate is at least $H(T_1,\dots,T_m)$, an excess of $\Delta_m=H(T_1,\dots,T_{m-1}\mid T_m)$ over the stage optimum $H(T_m)$, and one code attains this for all $m$ at once. The caption stages are therefore successively refinable if and only if every coarser caption is a function of the next finer one, $T_m=\phi_m(T_{m+1})$. When it is not, stage $m{+}1$ pays at least $\delta_m=H(T_m\mid T_{m+1})$.
\end{proposition}

The proof (\cref{app:proofs}) is the chain rule: a finer prefix contains every coarser one, so it must determine every coarser caption. The proposition is a rule about data. Caption levels should be produced by \emph{summarizing} the next finer level rather than written independently at each length; otherwise the finer stage pays for details that only the coarser caption mentions. \Cref{tab:ladder} tests the rule in our ablation setting: independently written levels raise the estimated gap $\hat\delta$ from 0.12 to 0.83 nats per token and cost 5 GenEval points and 31 points of nesting consistency. The condition is necessary for zero excess rate but does not say what the short levels should contain. Truncating one long caption satisfies it exactly and still costs 3 GenEval points, because the opening words of a long description are not a summary of it.

\subsection{The cost of each order}\label{sec:two-prices}

\begin{proposition}[Pixel-first can starve language]\label{prop:pixel-first}
For every $L$ there are a source $X$, a caption $T=\psi(X)$ with $H(T)>0$, and an ordered code that is successively refinable for $X$ under squared error at each of its $L$ stages, such that $I(Z_{\le k};T)=0$ for every $k\le L/2$.
\end{proposition}

Let $X=(A,B)$ with $A,B$ independent Gaussians, $\Var(A)\gg\Var(B)$, and $T=\ind[B>0]$. Reverse water-filling spends every nat below $\tfrac12\log(\Var A/\Var B)$ on $A$, so the first half of an optimal code is independent of $T$. The construction needs nothing unusual: in photographs the high-variance directions are illumination and low-frequency colour, which captions seldom mention.

Words-first moves the cost to the pixels. Let $T=T_M$; under the condition of \cref{prop:ladder} it determines every coarser level. The \emph{price} of putting words first at pixel distortion $D$ is the extra rate needed to determine $T$ and reach $D$, compared with a code that only reaches $D$:
\begin{equation}\label{eq:price}
\pi(D)\;=\;\min_{\E d(X,\hat X)\le D}I(X;T,\hat X)\;-\;R_X(D).
\end{equation}

\begin{proposition}[Words-first charges pixels, boundedly]\label{prop:price}
For every $D$, and any reconstruction $\hat X^*_D$ that achieves $R_X(D)$,
\[
\bigl[H(T)-R_X(D)\bigr]^+\;\le\;\pi(D)\;\le\;H\bigl(T\mid\hat X^*_D\bigr),
\]
and $\pi(D)=0$ if and only if some $\hat X^*_D$ determines $T$. If the pixel stages at $D_1>\dots>D_J$ are successively refinable, one words-first code attains the upper bound at every stage simultaneously.
\end{proposition}

\begin{corollary}[The price fades with fidelity]\label{cor:fade}
Suppose a reader $\tilde\psi$ and a margin $\rho>0$ satisfy $\tilde\psi(\hat x)=\psi(x)$ whenever $d(x,\hat x)\le\rho$, for all $x$ outside a set of probability $\epsilon$. Then for $D\le\rho(\tfrac12-\epsilon)$,
\[
\pi(D)\;\le\;h_b\bigl(\epsilon+D/\rho\bigr)+\bigl(\epsilon+D/\rho\bigr)\log|\cT|.
\]
If $\epsilon$ can be made arbitrarily small by shrinking $\rho$, that is, if captions are almost surely locally constant, then $\pi(D)\to0$ as $D\to0$.
\end{corollary}

\Cref{prop:price} locates the price. For short prefixes it is at least $H(T)-R_X(D)$, because any code that determines $T$ spends $H(T)$ nats on it. For long prefixes it is at most what a captioner cannot recover from the best reconstruction, which \cref{cor:fade} drives to zero once reconstructions can be captioned correctly. The corollary is qualitative for long captions, since $\log|\cT|$ is then large, so \cref{sec:price} measures the price directly. Both bounds can be estimated, which gives three predictions that \cref{sec:price} tests. \textbf{(P1)} A reconstruction-ordered prefix keeps little caption information. \textbf{(P2)} A words-first prefix is less faithful to the pixels at short lengths, and the gap closes with length. \textbf{(P3)} The measured price lies between the bounds and is small at full length.

\textbf{What the order buys.}\label{sec:buys-theory} With unlimited capacity every order induces the same $p(x\mid T)$; the order matters for how finite models with short samplers fail. In a words-first code at its stage optimum, the prefix up to the prompt's cut $k_T$ is a function of the prompt, $H(Z_{\le k_T}\mid T)=0$. An ideal sampler therefore makes no random choice about anything the prompt states, and a partial sample can be read back as a caption and checked against the prompt before later rungs build on it; a learned sampler only approximates this map, which is why the check of \cref{sec:correct} is useful. In a pixel-first code the prompt's content is spread over rungs that are sampled jointly with content the prompt never mentions. \Cref{sec:buys} tests these claims.

\section{\rung: a ladder from words to pixels}\label{sec:method}

\subsection{Building the ladder}\label{sec:ladder}

Let $h=\mathcal E_{\mathrm{vae}}(x)$. \rung{} encodes $h$ into an ordered code $z=(z_1,\dots,z_L)$ grouped into $L$ \emph{rungs} and reads it with two decoders: an image decoder that reconstructs $h$ from any prefix, and a language decoder $p_\psi$ that writes a caption from any prefix. Generation climbs the rungs and captioning descends them. Training uses, for every image, a caption ladder $T_1,\dots,T_M$ that satisfies \cref{prop:ladder} (\cref{sec:setup}).

Two terms train the code. The first makes every prefix decodable. The prefix length $k$ is drawn from a nested-dropout distribution $q$, and the image decoder is trained with flow matching \citep{lipman2023flow} on every truncation,
\begin{equation}\label{eq:rec}
\Lrec=\E_{k\sim q}\,\E_{t,\epsilon}\bigl\|v_\theta(h_t,t,z_{\le k})-(h-\epsilon)\bigr\|^2,\qquad h_t=t\,h+(1-t)\,\epsilon,
\end{equation}
so that any prefix decodes to a plausible image. The second term distinguishes \rung{} from a reconstruction-ordered tokenizer: each caption level must be recoverable from its matching prefix,
\begin{equation}\label{eq:lang}
\Llang=\sum_{m=1}^{M}-\log p_\psi\bigl(T_m\mid z_{\le k_m}\bigr).
\end{equation}
Gradients of $\Llang$ reach the encoder, so language, not variance, decides the competition for early rate. We minimize $\Lrec+\lambda\Llang$. Setting $\lambda=0$ gives nested dropout, larger $\lambda$ moves the code toward words-first, and we use $\lambda=1$ throughout (\cref{fig:dial}a traces the trade-off). A third term, introduced in \cref{sec:ascent}, trains the ascent model.

\textbf{Rate-matched cuts.}\label{sec:cuts} Rungs are quantized with FSQ \citep{mentzer2024fsq}, so the rate of a prefix can be measured. We set $\hR_k=\sum_{u}\hH(z_u)$, the sum of the empirical entropies of the code units $u$ in the first $k$ rungs; by subadditivity it upper-bounds the prefix entropy. The cut $k_m$ is the first code unit at which $\hR$ reaches $\hH(T_m)$, the caption's entropy under a small language model. This makes the stage optimum of \cref{prop:ladder} operational: level $m$ is read where the code has spent the rate that determining $T_m$ requires. The rates depend on the code being trained, so the cuts are recomputed every 5k steps during language anchoring (\cref{app:hparams}). In the trained code the five levels end at 1.2, 2.9, 6.0, 11 and 21\% of $\hR_L\approx2.5$k nats, inside rungs 1, 1, 2, 4 and 7 of 32. The captions occupy the first fifth of the code, and the remaining rungs hold what the captions leave free.

\subsection{Descent and ascent}\label{sec:ascent}

The two directions are not inverses of each other. Descent discards information while ascent has to supply it, so we give them separate operators instead of one invertible map (\cref{tab:design} tests the tied alternative).

\textbf{Descent} is deterministic: $x\mapsto z\mapsto z_{\le k}\mapsto$ greedy decoding of $p_\psi(\cdot\mid z_{\le k})$, one encoder pass and one short decode. Because every level reads a prefix of the same code, the 8-word caption agrees with the 45-word caption for a structural reason, whereas a prompted captioner can only be asked to agree (\cref{tab:caption}).

\textbf{Ascent} is stochastic. Given a prompt $T$, we locate its cut $k_T=\min\{k:\hR_k\ge\hH(T)\}$ and draw the code in eight blocks. The first four end at the ladder's cuts (rungs 1, 2, 4 and 7) and the last four cover rungs 8--11, 12--16, 17--23 and 24--32. Each block is drawn conditioned on the prompt and on the blocks before it,
\begin{equation}\label{eq:ascent}
z_{\le k+\Delta_b}\sim p_\theta\bigl(\cdot\mid z_{\le k},\,T,\,b\bigr),
\end{equation}
by a few-step conditional flow \citep{frans2025shortcut,geng2025meanflow} in the rung's pre-quantization space. At the stage optimum the blocks up to $k_T$ are a function of $T$ (\cref{sec:two-prices}), so the flow for these blocks learns a nearly deterministic map, and the later blocks carry the randomness. Sampling steps are allocated in proportion to each block's measured rate $\Delta\hR_b$, which is worth 3 GenEval points over a uniform allocation at the same budget (\cref{tab:design}). A consistency term couples the two directions without requiring invertibility: captions read back from generated prefixes must match the ladder of the training image, $\Lcons=\sum_m-\log p_\psi(T_m\mid\hat z_{\le k_m})$.

\subsection{Correcting errors where they appear}\label{sec:correct}

Each ladder block can be checked before the next one is drawn. We parse the prompt once into attribute tuples (object, count, colour, relation). After each of the first four blocks, an attribute head on the language decoder reads, directly from the prefix, the tuples stated by the levels that end in that block. The block is accepted if at least a fraction $\tau=0.8$ of these tuples agree with the prompt, where a tuple the prompt does not mention counts as agreeing; otherwise it is redrawn with fresh noise, up to $B=2$ times.

The purpose of the check is to localize the error. If the first block encodes three apples where the prompt asks for two, the wrong count is caught in that block, because the 8-word level states counts (\cref{app:laddercap}), and only that block is redrawn; best-of-$N$ would discard a whole sample for one wrong attribute. Blocks after the last cut are not checked, since no caption level describes them. The verifier is the model's own descent path, with no separate reward model, and it never sees the detectors that GenEval uses.

\textbf{Without training.}\label{sec:free} \emph{Editing} is a prefix swap: encode the image, replace the ladder prefix with one drawn from the edited caption, keep the suffix and decode. The prefix carries what the caption states and the suffix what it leaves free, so neither inversion nor attention manipulation is needed; \cref{app:qual} lists the cases in which the kept suffix does not fit the new prefix. \emph{Variation at a chosen granularity} is truncation: cut at rung $k$ and resample the rest, and the position of the cut decides what is preserved.

\section{Experiments}\label{sec:experiments}

\subsection{Setup}\label{sec:setup}

\textbf{Benchmarks.} For generation we use GenEval \citep{ghosh2023geneval}, DPG-Bench \citep{hu2024ella}, T2I-CompBench++ \citep{huang2025t2icompbench}, TIIF-Bench \citep{wei2025tiif}, WISE \citep{niu2025wise}, and FID and CMMD \citep{jayasumana2024cmmd} on MJHQ-30K. For understanding we use CapArena-Auto \citep{cheng2025caparena}, DCScore on DeCapBench \citep{ye2025decapbench} and CHAIR \citep{rohrbach2018chair}. We also report \textsc{GranBench}, 600 prompts whose attributes are annotated with the granularity at which they should appear, and \emph{nesting consistency} (NCC), the fraction of attribute tuples in a model's coarse caption that its own finer caption of the same image entails.

\textbf{Baselines.} We compare with generation-only models \citep{podell2024sdxl,esser2024sd3,flux2024,qin2025lumina2}, unified autoregressive models \citep{wang2024emu3,chen2025janus,xie2025showo2,deng2025bagel,wu2025omnigen2} and unified discrete-diffusion models \citep{yang2025mmada,shi2025muddit,xin2025dimoo}; with VLMs \citep{bai2025qwen25vl,zhu2025internvl3} as established references for captioning; and with two ordered codes, FlexTok \citep{bachmann2025flextok} and Semanticist \citep{wen2025semanticist}, each trained with the same text-to-image head (the \rung-L ascent model), data and capacity as \rung-L. \Cref{tab:order} separates the effect of the order from the rest of the tokenizer design. All baselines are evaluated using official model releases and standardized inference protocols without prompt rewriting.

\textbf{Implementation.} The ladder encoder is a ViT-B over SD3 VAE latents that produces $L=32$ rungs of 8 code units, quantized with FSQ (levels $[8,8,8,5,5,5]$, at most 11.1 nats per unit). The image decoder is a 24-layer DiT \citep{peebles2023dit} trained with flow matching, and the language decoder is Qwen3-1.7B \citep{yang2025qwen3} with LoRA adapters. The ascent model is an MMDiT \citep{esser2024sd3} at two sizes, \rung-L (1.3B) and \rung-XL (3.4B); the encoder and both decoders are shared, so the two sizes caption identically. Training has three stages, namely ladder pretraining, language anchoring on LadderCap, and ascent at $256^2$ and then $512^2$ (\cref{app:hparams}); \rung-XL takes about 9 days on 64 A100-80GB GPUs. Sampling uses 32 network evaluations, 24 for ascent and 8 for the image decoder, with guidance 1.5.

\textbf{LadderCap.} We build $M=5$ caption levels (${\approx}$8, 20, 45, 90 and 180 words) for 45M images drawn from Recap-DataComp \citep{li2024recap}, BLIP3-KALE \citep{awadalla2024kale}, PixelProse \citep{singla2024pixelprose}, DOCCI \citep{onoe2024docci}, ImageInWords \citep{garg2024iiw} and DCI \citep{urbanek2024dci}. Following \cref{prop:ladder}, the longest level is generated from the image (Qwen2.5-VL-7B, greedy) and every shorter level is a \emph{summary of the next longer level} (Qwen3-8B, greedy), never a fresh caption. The summarizer is instructed to keep a count or a relation in the 8-word level whenever the image contains more than one object. \Cref{app:laddercap} reports per-level statistics and an entailment audit on 5K images.

\subsection{Measuring the order and its price}\label{sec:price}

\begin{figure}[t]
\centering
\includegraphics[width=\linewidth]{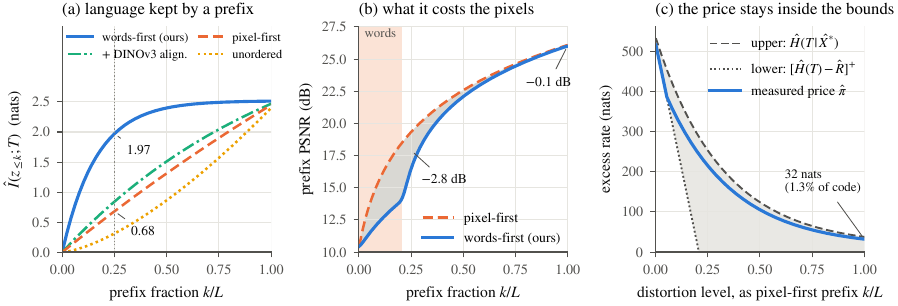}
\caption{\textbf{Empirical validation of rate--distortion properties.} Evaluated on RUNG-L ($\hat{R}_L \approx 2.5\text{k}$ nats). \textbf{(a)} Mutual information between code prefix and captions demonstrates that pixel-first codes discard semantic content at low rates. \textbf{(b)} Reconstruction fidelity shows words-first sacrifices 2.8\,dB PSNR within the language ladder, catching up to within 0.1\,dB at full code length. \textbf{(c)} Measured excess rate $\hat{\pi}(D)$ adheres strictly to the theoretical bounds of \cref{prop:price}, tapering to 73 nats (2.9\% of the code) at full fidelity.}
\label{fig:price}
\end{figure}

\begin{table}[t]
\centering
\caption{\textbf{Effect of visual code ordering signals.} Evaluated on RUNG-L ($256^2$) under identical capacity and data. $\hat{I}_{25}$ denotes the prefix--caption mutual information at one-quarter code length ($k/L = 1/4$). Quarter-code and full-code reconstructions evaluate image fidelity (PSNR, LPIPS) and distribution realism (rFID), alongside zero-shot text-to-image generation FID on MJHQ-30K.}
\label{tab:order}
\small
\setlength{\tabcolsep}{2.3pt}
\begin{tabular}{lccccccccc}
\toprule
& \multicolumn{3}{c}{Caption Retention \& Use} & \multicolumn{3}{c}{Quarter Code ($k/L=1/4$)} & \multicolumn{3}{c}{Full Code ($k/L=1$)} \\
\cmidrule(lr){2-4}\cmidrule(lr){5-7}\cmidrule(l){8-10}
Ordering Signal & $\hat{I}_{25}\uparrow$ & GenEval$\uparrow$ & NCC (\%) & PSNR$\uparrow$ & LPIPS$\downarrow$ & rFID$\downarrow$ & PSNR$\uparrow$ & rFID$\downarrow$ & Gen.\,FID$\downarrow$ \\
\midrule
Unordered (Vanilla)           & 0.31 & 0.71 & 22.1 & --   & --   & --  & 26.2 & 0.70 & 6.8 \\
Pixel-First (Nested Dropout)  & 0.68 & 0.76 & 41.5 & 19.3 & 0.36 & 2.3 & 26.1 & 0.71 & 6.3 \\
DINOv3 Feature Alignment     & 0.84 & 0.78 & 48.3 & 18.8 & 0.38 & 2.2 & 26.0 & 0.72 & 6.3 \\
\rowcolor{rungrow}Words-First (RUNG-L) & \textbf{1.97} & \textbf{0.83} & \textbf{94.1} & 16.5 & 0.45 & 2.1 & 26.0 & 0.74 & \textbf{6.2} \\
\bottomrule
\end{tabular}
\end{table}

\Cref{tab:order} and \cref{fig:price,fig:dial}a retrain the tokenizer and the \rung-L ascent model at full scale for each ordering signal, with everything else fixed. $\hat I$ is a Barber--Agakov lower bound \citep{barber2003im} on the information between a prefix and the finest caption $T=T_5$, reported in nats per caption token ($\hH(T_5)/|T_5|=2.5$); an InfoNCE bound \citep{poole2019bounds} ranks the codes identically (\cref{app:estimators}). The measured price $\hat\pi(D)$ is the rate at which the words-first code has both passed its last cut \emph{and} reached PSNR $D$, minus the rate at which the pixel-first code reaches $D$. The upper bound is the cross-entropy of a captioning probe on pixel-first reconstructions (\cref{app:estimators}).

\textbf{P1: reconstruction order starves language.} At a quarter of the code the reconstruction-ordered prefix keeps 0.68 nats of caption information per token and the language-ordered prefix 1.97 (\cref{fig:price}a). Aligning the encoder to DINOv3 features \citep{simeoni2025dinov3} in the style of REPA \citep{yu2025repa} raises the first number to 0.84, so visual self-supervision moves the order toward language, but only a small part of the way. At full length every code determines the image, and the four estimates meet at the probe's ceiling of 2.2 nats per token. This indicates that the probe measures information rather than how easily a particular code can be read.

\textbf{P2: words-first costs pixel fidelity where the words are.} The words-first prefix is 2.8\,dB less faithful at a quarter of the code (\cref{tab:order}), and the gap is largest inside the ladder, while the code is still encoding captions (\cref{fig:price}b). The cost falls on fidelity to the particular image, and realism is unaffected: because the decoder samples, quarter-code decodes remain realistic (rFID 2.1 against 2.3; cf.\ \citealp{blau2019rdp}). The gap closes with length, to 0.5\,dB at half the code and 0.1\,dB at full length.

\textbf{P3: the price stays inside the bounds and is small at full length.} Measured in rate (\cref{fig:price}c), the price equals the lower bound $[\hH(T)-\hR]^+$ while the words-first code is still inside its ladder, which holds by construction (\cref{app:estimators}), and it stays below the estimated upper bound $\hH(T\mid\hat X^*)$ everywhere (207 against 394 nats at a quarter). It ends at 73 nats, 2.9\% of the code. Full-length reconstruction (rFID 0.74 against 0.71) and text-to-image FID (6.2 against 6.3) differ by less than the variation across seeds (0.03 and 0.1). Neither application uses the pixels of a short prefix: captioning reads the prefix for its words, and generation decodes the full code.

\textbf{The order is a dial.} \Cref{fig:dial}a sweeps $\lambda$. Each step toward language buys caption information and GenEval at the expense of quarter-code PSNR. GenEval saturates at $\lambda{=}1$, and at $\lambda{=}3$ the code gains little more caption information and starts to lose fidelity at full length (PSNR 25.7). DINOv3 alignment lies inside the trade-off curve, so it buys caption information at a worse rate of exchange than supervision with language. With everything else matched, the order accounts for 7 of the 16 GenEval points between \rung-L and FlexTok with the same head (0.76 to 0.83 in \cref{tab:order}); the other 9 come from the rest of the design, which \cref{tab:design} breaks down in the ablation setting.

\subsection{What the order buys}\label{sec:buys}

\begin{table}[t]
\centering
\caption{\textbf{Fine-grained visual compositionality on GenEval.} Evaluated under the standard GenEval suite across object count, color binding, and spatial positioning without prompt expansion. \#Par denotes generator parameters during sampling. Lumina-DiMOO incorporates template-level GRPO alignment during post-training.}
\label{tab:geneval}
\small
\setlength{\tabcolsep}{4.45pt}
\begin{tabular}{llccccccccc}
\toprule
Paradigm & Model & \#Par & Single & Two & Count & Colors & Position & Attribute & Overall \\
\midrule
\multirow{4}{*}{Generation}
& SDXL               & 2.6B & 0.98 & 0.74 & 0.39 & 0.85 & 0.15 & 0.23 & 0.55 \\
& SD3-Medium         & 2.0B & 0.99 & 0.94 & 0.72 & 0.89 & 0.33 & 0.60 & 0.74 \\
& Lumina-Image 2.0   & 2.6B & 0.94 & 0.87 & 0.67 & 0.89 & 0.62 & 0.62 & 0.73 \\
& FLUX.1-dev         & 12B  & 0.98 & 0.81 & 0.74 & 0.79 & 0.22 & 0.45 & 0.66 \\
\midrule
\multirow{5}{*}{Unified AR}
& Emu3-Gen           & 8.0B & 0.98 & 0.71 & 0.34 & 0.81 & 0.17 & 0.21 & 0.54 \\
& Show-o2            & 7.0B & 1.00 & 0.87 & 0.58 & \textbf{0.92} & 0.52 & 0.62 & 0.76 \\
& Janus-Pro          & 7.0B & 0.99 & 0.89 & 0.59 & 0.90 & 0.79 & 0.66 & 0.80 \\
& OmniGen2           & 7.0B & 1.00 & \textbf{0.95} & 0.64 & 0.88 & 0.55 & 0.76 & 0.80 \\
& BAGEL              & 14B  & 0.99 & 0.94 & 0.81 & 0.88 & 0.64 & 0.63 & 0.82 \\
\midrule
\multirow{3}{*}{Discrete Diff.}
& Muddit             & 1.0B & 0.98 & 0.72 & 0.54 & 0.82 & 0.19 & 0.41 & 0.61 \\
& MMaDA              & 8.0B & 0.99 & 0.76 & 0.61 & 0.84 & 0.20 & 0.37 & 0.63 \\
& Lumina-DiMOO       & 8.0B & 1.00 & 0.94 & 0.85 & 0.89 & 0.85 & 0.76 & 0.88 \\
\midrule
\rowcolor{baserow}& Semanticist + T2I Head & 1.3B & 0.98 & 0.80 & 0.49 & 0.85 & 0.33 & 0.45 & 0.65 \\
\rowcolor{baserow}& FlexTok + T2I Head     & 1.3B & 0.98 & 0.82 & 0.51 & 0.86 & 0.36 & 0.47 & 0.67 \\
\midrule
\rowcolor{rungrow}& RUNG-L                 & 1.3B & 0.99 & 0.89 & 0.78 & 0.87 & 0.75 & 0.72 & 0.83 \\
\rowcolor{rungrow}& RUNG-XL                & 3.4B & 0.99 & 0.93 & 0.85 & 0.90 & 0.82 & 0.79 & 0.88 \\
\rowcolor{rungrow}& RUNG-XL + Corr.        & 3.4B & \textbf{1.00} & 0.94 & \textbf{0.89} & \textbf{0.92} & \textbf{0.87} & \textbf{0.84} & \textbf{0.91} \\
\bottomrule
\end{tabular}
\end{table}

\begin{figure}[t]
\centering
\includegraphics[width=0.84\linewidth]{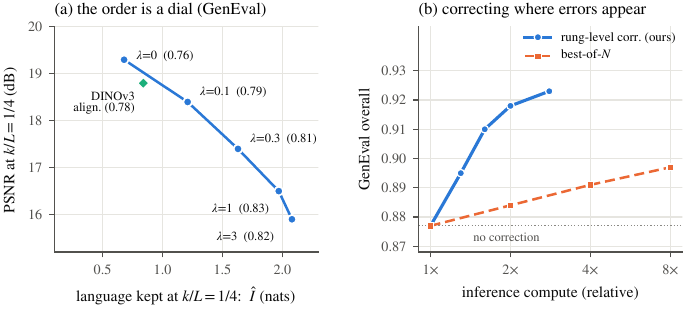}
\caption{\textbf{Left: the order is a dial.} Sweeping $\lambda$ trades quarter-code PSNR for caption information (GenEval in parentheses); $\lambda{=}0$ is nested dropout, and DINOv3 alignment lies inside the trade-off curve. \textbf{Right: test-time compute.} Rung-level correction (redraw budgets $B=1$ to $4$; \cref{app:hparams}) against best-of-$N$ with the same verifier, in network evaluations relative to one uncorrected sample.}
\label{fig:dial}
\end{figure}

\textbf{Generation.} \Cref{tab:geneval,tab:dpg} give the main comparison. \rung-XL ties Lumina-DiMOO at 0.88 GenEval with half its sampling-time parameters (4.0B including the image decoder, against 8B; \cref{app:hparams}) and half its network evaluations, and without reward tuning. Rung-level correction raises \rung-XL to 0.91. Per category, \rung{} does \emph{not} lead everywhere. On two-object prompts it trails OmniGen2, SD3-Medium, BAGEL and Lumina-DiMOO, and on colour it trails Show-o2, as one would expect from a smaller generator on properties that improve with capacity. Against the models without reward tuning, \rung-XL gains in counting ($+0.04$ over BAGEL), position ($+0.03$ over Janus-Pro) and attribute binding ($+0.03$ over OmniGen2). A short prompt states these three properties explicitly, and the words-first order places them in the prompt's prefix (\cref{sec:buys-theory}). Lumina-DiMOO is ahead of \rung-XL on position (0.85 against 0.82) and level on counting; with correction \rung-XL exceeds every baseline on all three (0.89, 0.87 and 0.84). T2I-CompBench++ shows the same pattern: the gain over Lumina-DiMOO is concentrated in numeracy ($+0.052$) and 2D spatial relations ($+0.048$) and is close to zero for non-spatial relations, which short captions rarely pin down. Extra compute is better spent on rungs than on samples: correction reaches 0.91 at $1.2\times$ the network evaluations and best-of-$N$ with the same verifier 0.89 at $4\times$ (\cref{fig:dial}b), since a redraw keeps the blocks that already passed.

\begin{table}[!htbp]
\centering
\caption{\textbf{Instruction following, compositional synthesis, and efficiency.} Benchmarked across GenEval, DPG-Bench, T2I-CompBench++ (T2I-C++), TIIF-Bench, WISE, and MJHQ-30K FID. Total NFE aggregates all sampling network evaluations (generator, latent flow decoder, and verifier passes); CFG forward passes are batched. Latency is measured at $512^2$ resolution with batch size 1 on an NVIDIA A100 GPU.}
\label{tab:dpg}
\small
\setlength{\tabcolsep}{5.2pt}
\begin{tabular}{lcccccccc}
\toprule
Model & GenEval & DPG & T2I-C++ & TIIF & WISE & FID$\downarrow$ & Total NFE$\downarrow$ & Latency (s)$\downarrow$ \\
\midrule
SDXL                 & 0.55 & 74.7 & 0.408 & 51.2 & 0.43 & 8.8 & 50  & 3.1 \\
FLUX.1-dev           & 0.66 & 83.8 & 0.502 & 72.6 & 0.50 & 6.1 & 50  & 4.8 \\
Lumina-Image 2.0     & 0.73 & 87.2 & 0.487 & 69.1 & 0.43 & 6.8 & 40  & 3.4 \\
Janus-Pro            & 0.80 & 84.2 & 0.471 & 66.3 & 0.35 & 8.0 & 576 & 9.1 \\
BAGEL                & 0.82 & 85.1 & 0.524 & 71.4 & 0.52 & 6.4 & 50  & 7.2 \\
Lumina-DiMOO         & 0.88 & 86.0 & 0.551 & 74.8 & 0.60 & 6.9 & 64  & 6.3 \\
\midrule
\rowcolor{baserow}FlexTok + T2I Head   & 0.67 & 80.3 & 0.421 & 58.7 & 0.34 & 7.9 & 32  & 1.9 \\
\midrule
\rowcolor{rungrow}RUNG-L               & 0.83 & 85.3 & 0.538 & 72.9 & 0.54 & 6.2 & \textbf{32}  & \textbf{1.4} \\
\rowcolor{rungrow}RUNG-XL              & 0.88 & 87.6 & 0.581 & 76.2 & 0.61 & 5.8 & \textbf{32}  & 2.1 \\
\rowcolor{rungrow}RUNG-XL + Corr.      & \textbf{0.91} & \textbf{88.4} & \textbf{0.607} & \textbf{78.5} & \textbf{0.63} & \textbf{5.6} & 38  & 2.5 \\
\bottomrule
\end{tabular}
\end{table}

\begin{table}[!htbp]

\centering
\caption{\textbf{Multimodal captioning and structural nesting consistency (NCC).} Evaluated on CapArena-Auto (CapA., $[-100, 100]$), DeCapBench DCScore (DCS, $[0, 100]$), CHAIR$_s$ ($[0, 100]$, lower is better), and \textsc{GranBench}. NCC measures the fraction of attribute tuples in a model's coarse caption entailed by its fine caption on identical images.}
\label{tab:caption}
\small
\setlength{\tabcolsep}{7.5pt}
\begin{tabular}{lccccc}
\toprule
Model & CapA. & DCS & CHAIR$_s\!\downarrow$ & NCC (\%) & GranBench \\
\midrule
Qwen2.5-VL-7B (Reference) & $-12.3$ & 42.8 & 11.2 & 61.4 & 44.7 \\
InternVL3-8B (Reference)  & $-18.7$ & 39.5 & 13.5 & 57.9 & 41.2 \\
BAGEL                     & $-46.5$ & 31.2 & 18.4 & 48.2 & 33.6 \\
Lumina-DiMOO              & $-53.2$ & 29.8 & 21.0 & 45.1 & 31.9 \\
\midrule
\rowcolor{rungrow}\rung{} Descent & \textbf{-24.9} & \textbf{38.6} & \textbf{12.8} & \textbf{94.1} & \textbf{71.3} \\
\bottomrule
\end{tabular}
\end{table}

\textbf{Understanding.} With a 1.7B language decoder, descent trails the 7B VLMs on detail (CapArena-Auto $-24.9$ against $-12.3$ and $-18.7$), lies between them on hallucination, and is ahead of both unified models on every column (\cref{tab:caption}). On nesting it differs from all prompted models, 94.1\% against 45--61\%, because its two captions read two prefixes of one code instead of coming from two independent decodes. Most of the remaining 5.9\% comes from the language decoder: attribute tuples read directly from the prefixes nest 99.2\% of the time.

\textbf{Editing and variation.} On PIE-Bench \citep{ju2024pnp}, a prefix swap reaches CLIP similarity 26.4 on the whole image and 23.1 on the edited region at a structure distance of $19.6\times10^{-3}$, against 25.8/22.7 at 27.4 for BAGEL's editing mode and 25.2/22.2 at 40.6 for RF-Inversion \citep{rout2025rfinversion}. The swap costs one encode and one decode; inversion costs 2--3$\times$ more. A cut at rung 8, one rung past the finest caption (${\approx}625$ nats), keeps 96\% of the attribute tuples of the 180-word description; a cut at rung 20 (${\approx}1.56$k nats) also keeps layout and texture and mainly changes lighting (\cref{app:qual}).

\textbf{Transfer.} Radiology reports come with a two-level ladder, since the impression approximately summarizes the findings. On MIMIC-CXR \citep{johnson2019mimic}, \rung{} reaches 0.331 RadGraph-F1 \citep{jain2021radgraph} and 0.495 CheXbert-F1 \citep{smit2020chexbert} with 0.91 two-level consistency, against 0.312/0.478 for a matched baseline that writes the two sections independently; the result concerns structure only, not clinical accuracy.

\begin{table}[!htbp]

\centering
\caption{\textbf{Validation of Proposition 1 on caption hierarchy construction.} Evaluated on \rung-L ($256^2$, 12M data). $\hat\delta = \hat H(T_m \mid T_{m+1})$ quantifies inter-stage information leakage in nats per token. Violating the Markov summarization condition imposes severe rate penalties and degrades semantic nesting consistency.}
\label{tab:ladder}
\small
\setlength{\tabcolsep}{7.5pt}
\begin{tabular}{lccc}
\toprule
Construction & $\hat\delta$ & GenEval & NCC (\%)\\
\midrule
Independent per length & 0.83 & 0.74 & 61.8\\
Truncated long caption & \textbf{0.04} & 0.76 & 90.4\\
\rowcolor{rungrow}Successive summary & 0.12 & \textbf{0.79} & \textbf{92.7}\\
\bottomrule
\end{tabular}
\end{table}

\textbf{Ablations}\label{sec:ablations} use \rung-L at $256^2$ on a 12M-image subset (default GenEval 0.79, against 0.83 at full scale). \emph{The data rule matters as much as the loss} (\cref{tab:ladder}). Independently written levels raise $\hat\delta$ from 0.12 to 0.83 and cost 5 GenEval points, as \cref{prop:ladder} predicts. Truncating one long caption satisfies the chain condition but loses 3 points, because its short levels open a long description without summarizing it (\cref{sec:caption-stages}). \emph{Design} (\cref{app:design}). Performance saturates at $L=32$. Per-rung few-step flows beat autoregressive and single-diffusion ascent at a third and a ninth of their latency. Frozen text embeddings cost 9 to 12 points, and tying the two directions into one time-reversible operator costs 4.

\FloatBarrier
\section{Limitations and conclusion}\label{sec:conclusion}

\textbf{Limitations.} The price is real: a words-first prefix is a worse picture of the particular image than a pixel-first prefix of the same length, so applications that need faithful pixels from few bits should use a small $\lambda$ or none. The theory assumes deterministic captions and informational rates, \cref{cor:fade} is qualitative for long captions, and our checks rely on estimators that are themselves bounds. LadderCap inherits biases shared by the language models that build it. Published baseline protocols differ in resolution and prompt handling. We train at $512^2$ and ran no human preference study, so we claim nothing about image quality against generation-only systems at $1024^2$.

\textbf{Conclusion.} An ordered code always decides what its first bits describe. With a reconstruction loss the decision is made by variance, and \cref{prop:pixel-first} shows that it can leave language with nothing. We made the opposite choice explicit, with a distortion schedule that fixes each stage's optimum, a condition under which the caption stages cost nothing, and a bounded cost in pixels paid at short prefixes. In exchange, generation extends one code and captioning truncates it. Choosing the schedule per image, deciding where the words end and the pixels begin, is left to future work.

\subsection*{Reproducibility statement}
The ladder objectives are in \cref{sec:ladder}, ascent and self-correction in \cref{sec:ascent,sec:correct}, and complete proofs of \cref{prop:ladder,prop:pixel-first,prop:price} and \cref{cor:fade}, with all assumptions, in \cref{app:proofs}. The estimators for prefix information, rate and the price bounds are specified in \cref{app:estimators}. Architecture, staging and hardware are in \cref{sec:setup} and all hyperparameters in \cref{app:hparams}. LadderCap uses public datasets and public models; prompts, the entailment audit and per-level statistics are in \cref{app:laddercap}. Code, the LadderCap pipeline, \textsc{GranBench}, the NCC evaluator and the scripts that produce \cref{fig:price} will be made available.

\subsection*{Ethics statement}
\rung{} inherits the risks of its class: representational bias absorbed from web-scale caption data and potential misuse for deceptive imagery. Control over granularity adds no new capability here, although the prefix-swap edit makes targeted semantic edits cheaper, which matters for provenance tooling. The MIMIC-CXR experiment uses a credentialed dataset under its data use agreement and reports a structural transfer result; nothing in it is validated for clinical use.

\subsection*{Author contributions}
R. Gong and H. Gong conceived the core theoretical formulation, designed the cross-modal distortion schedule, developed the \rung{} architecture, and executed the experimental pipeline. Both authors contributed equally to mathematical proofs, empirical evaluations, and manuscript composition.

\bibliography{refs}

@article{albergo2023,
  title={Stochastic interpolants: A unifying framework for flows and diffusions},
  author={Albergo, Michael S. and Boffi, Nicholas M. and Vanden-Eijnden, Eric},
  journal={arXiv preprint arXiv:2303.08797}, year={2023}}

@article{awadalla2024kale,
  title={{BLIP3-KALE}: Knowledge augmented large-scale dense captions},
  author={Awadalla, Anas and Xue, Le and Shu, Manli and Yan, An and Lee, Jun and Nguyen, Thao and Zhu, Mohamed and Yang, Yuhui and Bansal, Mohit and Schmidt, Ludwig and others},
  journal={arXiv preprint arXiv:2411.07461}, year={2024}}

@inproceedings{bachmann2025flextok,
  title={{FlexTok}: Resampling images into {1D} token sequences of flexible length},
  author={Bachmann, Roman and Allardice, Jesse and Mizrahi, David and Fini, Enrico and Kar, O{\u{g}}uzhan Fatih and Amirloo, Elmira and El-Nouby, Alaaeldin and Zamir, Amir and Dehghan, Afshin},
  booktitle={ICML}, year={2025}}

@article{bai2025qwen25vl,
  title={{Qwen2.5-VL} technical report},
  author={Bai, Shuai and Chen, Keqin and Liu, Xuejing and Wang, Jialin and Ge, Wenbin and Song, Sibo and Dang, Kai and Wang, Peng and Wang, Shijie and Tang, Jun and others},
  journal={arXiv preprint arXiv:2502.13923}, year={2025}}

@misc{flux2024,
  title={{FLUX.1}}, author={{Black Forest Labs}}, year={2024},
  howpublished={\url{https://github.com/black-forest-labs/flux}}}

@inproceedings{chang2023muse,
  title={Muse: Text-to-image generation via masked generative transformers},
  author={Chang, Huiwen and Zhang, Han and Barber, Jarred and Maschinot, AJ and Lezama, Jos{\'e} and Jiang, Lu and Yang, Ming-Hsuan and Murphy, Kevin and Freeman, William T. and Rubinstein, Michael and others},
  booktitle={ICML}, year={2023}}

@article{chen2025janus,
  title={{Janus-Pro}: Unified multimodal understanding and generation with data and model scaling},
  author={Chen, Xiaokang and Wu, Zhiyu and Liu, Xingchao and Pan, Zizheng and Liu, Wen and Xie, Zhenda and Yu, Xingkai and Ruan, Chong},
  journal={arXiv preprint arXiv:2501.17811}, year={2025}}

@inproceedings{cheng2025caparena,
  title={{CapArena}: Benchmarking and analyzing detailed image captioning in the {LLM} era},
  author={Cheng, Kanzhi and Song, Wenpo and Fan, Jiaxin and Ma, Zheng and Sun, Qiushi and Xu, Fangzhi and Yan, Chenyang and Chen, Nuo and Zhang, Jianbing and Chen, Jiajun},
  booktitle={Findings of ACL}, year={2025}}

@article{deng2025bagel,
  title={Emerging properties in unified multimodal pretraining},
  author={Deng, Chaorui and Zhu, Deyao and Li, Kunchang and Gou, Chenhui and Li, Feng and Wang, Zeyu and Zhong, Shu and Yu, Weihao and Nie, Xiaonan and Song, Ziang and others},
  journal={arXiv preprint arXiv:2505.14683}, year={2025}}

@article{equitz1991,
  title={Successive refinement of information},
  author={Equitz, William H. R. and Cover, Thomas M.},
  journal={IEEE Transactions on Information Theory}, volume={37}, number={2}, pages={269--275}, year={1991}}

@inproceedings{esser2024sd3,
  title={Scaling rectified flow transformers for high-resolution image synthesis},
  author={Esser, Patrick and Kulal, Sumith and Blattmann, Andreas and Entezari, Rahim and M{\"u}ller, Jonas and Saini, Harry and Levi, Yam and Lorenz, Dominik and Sauer, Axel and Boesel, Frederic and others},
  booktitle={ICML}, year={2024}}

@inproceedings{frans2025shortcut,
  title={One step diffusion via shortcut models},
  author={Frans, Kevin and Hafner, Danijar and Levine, Sergey and Abbeel, Pieter},
  booktitle={ICLR}, year={2025}}

@inproceedings{garg2024iiw,
  title={{ImageInWords}: Unlocking hyper-detailed image descriptions},
  author={Garg, Roopal and Burns, Andrea and Ayan, Burcu Karagol and Bitton, Yonatan and Montgomery, Ceslee and Onoe, Yasumasa and Bunner, Andrew and Krishna, Ranjay and Baldridge, Jason and Soricut, Radu},
  booktitle={EMNLP}, year={2024}}

@inproceedings{geng2025meanflow,
  title={Mean flows for one-step generative modeling},
  author={Geng, Zhengyang and Deng, Mingyang and Bai, Xingjian and Kolter, J. Zico and He, Kaiming},
  booktitle={NeurIPS}, year={2025}}

@inproceedings{ghosh2023geneval,
  title={{GenEval}: An object-focused framework for evaluating text-to-image alignment},
  author={Ghosh, Dhruba and Hajishirzi, Hannaneh and Schmidt, Ludwig},
  booktitle={NeurIPS Datasets and Benchmarks}, year={2023}}

@inproceedings{han2025infinity,
  title={Infinity: Scaling bitwise autoregressive modeling for high-resolution image synthesis},
  author={Han, Jian and Liu, Jinlai and Jiang, Yi and Yan, Bin and Zhang, Yuqi and Yuan, Zehuan and Peng, Bingyue and Liu, Xiaobing},
  booktitle={CVPR}, year={2025}}

@article{he2025flowtok,
  title={{FlowTok}: Flowing seamlessly across text and image tokens},
  author={He, Ju and Yu, Qihang and Shen, Qihao and Yuille, Alan and Chen, Liang-Chieh},
  journal={arXiv preprint arXiv:2503.10772}, year={2025}}

@article{hu2024ella,
  title={{ELLA}: Equip diffusion models with {LLM} for enhanced semantic alignment},
  author={Hu, Xiwei and Wang, Rui and Fang, Yixiao and Fu, Bin and Cheng, Pei and Yu, Gang},
  journal={arXiv preprint arXiv:2403.05135}, year={2024}}

@article{huang2025t2icompbench,
  title={{T2I-CompBench++}: An enhanced and comprehensive benchmark for compositional text-to-image generation},
  author={Huang, Kaiyi and Duan, Chengqi and Sun, Kaiyue and Xie, Enze and Li, Zhenguo and Liu, Xihui},
  journal={IEEE Transactions on Pattern Analysis and Machine Intelligence}, year={2025}}

@inproceedings{jain2021radgraph,
  title={{RadGraph}: Extracting clinical entities and relations from radiology reports},
  author={Jain, Saahil and Agrawal, Ashwin and Saporta, Adriel and Truong, Steven Q. H. and Duong, Du Nguyen and Bui, Tan and Chambon, Pierre and Zhang, Yuhao and Lungren, Matthew P. and Ng, Andrew Y. and others},
  booktitle={NeurIPS Datasets and Benchmarks}, year={2021}}

@inproceedings{jayasumana2024cmmd,
  title={Rethinking {FID}: Towards a better evaluation metric for image generation},
  author={Jayasumana, Sadeep and Ramalingam, Srikumar and Veit, Andreas and Glasner, Daniel and Chakrabarti, Ayan and Kumar, Sanjiv},
  booktitle={CVPR}, year={2024}}

@article{johnson2019mimic,
  title={{MIMIC-CXR-JPG}: A large publicly available database of labeled chest radiographs},
  author={Johnson, Alistair E. W. and Pollard, Tom J. and Greenbaum, Nathaniel R. and Lungren, Matthew P. and Deng, Chih-ying and Peng, Yifan and Lu, Zhiyong and Mark, Roger G. and Berkowitz, Seth J. and Horng, Steven},
  journal={arXiv preprint arXiv:1901.07042}, year={2019}}

@inproceedings{kusupati2022matryoshka,
  title={Matryoshka representation learning},
  author={Kusupati, Aditya and Bhatt, Gantavya and Rege, Aniket and Wallingford, Matthew and Sinha, Aditya and Ramanujan, Vivek and Howard-Snyder, William and Chen, Kaifeng and Kakade, Sham and Jain, Prateek and Farhadi, Ali},
  booktitle={NeurIPS}, year={2022}}

@article{li2025lavidao,
  title={{Lavida-O}: Elastic large masked diffusion models for unified multimodal understanding and generation},
  author={Li, Shufan and Gu, Jiuxiang and Liu, Kangning and Lin, Zhe and Wei, Zijun and Grover, Aditya and Kuen, Jason},
  journal={arXiv preprint arXiv:2509.19244}, year={2025}}

@inproceedings{lipman2023flow,
  title={Flow matching for generative modeling},
  author={Lipman, Yaron and Chen, Ricky T. Q. and Ben-Hamu, Heli and Nickel, Maximilian and Le, Matthew},
  booktitle={ICLR}, year={2023}}

@inproceedings{liu2025crossflow,
  title={Flowing from words to pixels: A framework for cross-modality evolution},
  author={Liu, Qihao and Liu, Xi and Zhou, Yizhou and Hou, Jie and Yuille, Alan and Zhang, Liang-Chieh},
  booktitle={CVPR}, year={2025}}

@inproceedings{mentzer2024fsq,
  title={Finite scalar quantization: {VQ-VAE} made simple},
  author={Mentzer, Fabian and Minnen, David and Agustsson, Eirikur and Tschannen, Michael},
  booktitle={ICLR}, year={2024}}

@article{miwa2025onedpiece,
  title={{One-D-Piece}: Image tokenizer meets quality-controllable compression},
  author={Miwa, Keita and Sasaki, Kento and Arai, Hidehisa and Takahashi, Tsubasa and Yamaguchi, Yu},
  journal={arXiv preprint arXiv:2501.10064}, year={2025}}

@article{niu2025wise,
  title={{WISE}: A world knowledge-informed semantic evaluation for text-to-image generation},
  author={Niu, Yuwei and Ning, Munan and Zheng, Mengren and Lin, Bin and Jin, Peng and Liao, Jiaqi and Ning, Kunpeng and Zhu, Bin and Yuan, Li},
  journal={arXiv preprint arXiv:2503.07265}, year={2025}}

@inproceedings{onoe2024docci,
  title={{DOCCI}: Descriptions of connected and contrasting images},
  author={Onoe, Yasumasa and Rane, Sunayana and Berger, Zachary and Bitton, Yonatan and Cho, Jaemin and Garg, Roopal and Ku, Alexander and Parekh, Zarana and Pont-Tuset, Jordi and Tanzer, Garrett and Wang, Su and Baldridge, Jason},
  booktitle={ECCV}, year={2024}}

@inproceedings{peebles2023dit,
  title={Scalable diffusion models with transformers},
  author={Peebles, William and Xie, Saining},
  booktitle={ICCV}, year={2023}}

@inproceedings{podell2024sdxl,
  title={{SDXL}: Improving latent diffusion models for high-resolution image synthesis},
  author={Podell, Dustin and English, Zion and Lacey, Kyle and Blattmann, Andreas and Dockhorn, Tim and M{\"u}ller, Jonas and Penna, Joe and Rombach, Robin},
  booktitle={ICLR}, year={2024}}

@inproceedings{poole2019bounds,
  title={On variational bounds of mutual information},
  author={Poole, Ben and Ozair, Sherjil and van den Oord, Aaron and Alemi, Alexander A. and Tucker, George},
  booktitle={ICML}, year={2019}}

@article{qin2025lumina2,
  title={{Lumina-Image 2.0}: A unified and efficient image generative framework},
  author={Qin, Qi and Zhuo, Le and Xin, Yi and Du, Ruoyi and Li, Zhen and Fu, Bin and Lu, Yiting and Yuan, Jiakang and Li, Xinyue and Liu, Dongyang and others},
  journal={arXiv preprint arXiv:2503.21758}, year={2025}}

@article{ramesh2022dalle2,
  title={Hierarchical text-conditional image generation with {CLIP} latents},
  author={Ramesh, Aditya and Dhariwal, Prafulla and Nichol, Alex and Chu, Casey and Chen, Mark},
  journal={arXiv preprint arXiv:2204.06125}, year={2022}}

@article{rimoldi1994,
  title={Successive refinement of information: Characterization of the achievable rates},
  author={Rimoldi, Bixio},
  journal={IEEE Transactions on Information Theory}, volume={40}, number={1}, pages={253--259}, year={1994}}

@inproceedings{rippel2014nested,
  title={Learning ordered representations with nested dropout},
  author={Rippel, Oren and Gelbart, Michael A. and Adams, Ryan P.},
  booktitle={ICML}, year={2014}}

@inproceedings{rohrbach2018chair,
  title={Object hallucination in image captioning},
  author={Rohrbach, Anna and Hendricks, Lisa Anne and Burns, Kaylee and Darrell, Trevor and Saenko, Kate},
  booktitle={EMNLP}, year={2018}}

@inproceedings{rout2025rfinversion,
  title={Semantic image inversion and editing using rectified stochastic differential equations},
  author={Rout, Litu and Chen, Yujia and Ruiz, Nataniel and Caramanis, Constantine and Shakkottai, Sanjay and Chu, Wen-Sheng},
  booktitle={ICLR}, year={2025}}

@article{shi2025muddit,
  title={Muddit: Liberating generation beyond text-to-image with a unified discrete diffusion model},
  author={Shi, Qingyu and Bai, Jinbin and Zhao, Zhuoran and Chai, Wenhao and Yu, Kaidong and Wu, Jianzong and Song, Shuangyong and Tong, Yunhai and Li, Xiangtai and Li, Xuelong and Yan, Shuicheng},
  journal={arXiv preprint arXiv:2505.23606}, year={2025}}

@article{simeoni2025dinov3,
  title={{DINOv3}},
  author={Sim{\'e}oni, Oriane and Vo, Huy V. and Seitzer, Maximilian and Baldassarre, Federico and Oquab, Maxime and Jose, Cijo and Khalidov, Vasil and Szafraniec, Marc and Yi, Seungeun and others},
  journal={arXiv preprint arXiv:2508.10104}, year={2025}}

@article{singla2024pixelprose,
  title={From pixels to prose: A large dataset of dense image captions},
  author={Singla, Vasu and Yue, Kaiyu and Paul, Sukriti and Shirkavand, Reza and Jayawardhana, Mayuka and Ganjdanesh, Alireza and Huang, Heng and Bhatele, Abhinav and Somepalli, Gowthami and Goldstein, Tom},
  journal={arXiv preprint arXiv:2406.10328}, year={2024}}

@inproceedings{smit2020chexbert,
  title={Combining automatic labelers and expert annotations for accurate radiology report labeling using {BERT}},
  author={Smit, Akshay and Jain, Saahil and Rajpurkar, Pranav and Pareek, Anuj and Ng, Andrew Y. and Lungren, Matthew P.},
  booktitle={EMNLP}, year={2020}}

@inproceedings{tian2024var,
  title={Visual autoregressive modeling: Scalable image generation via next-scale prediction},
  author={Tian, Keyu and Jiang, Yi and Yuan, Zehuan and Peng, Bingyue and Wang, Liwei},
  booktitle={NeurIPS}, year={2024}}

@inproceedings{urbanek2024dci,
  title={A picture is worth more than 77 text tokens: Evaluating {CLIP}-style models on dense captions},
  author={Urbanek, Jack and Bordes, Florian and Astolfi, Pietro and Williamson, Mary and Sharma, Vasu and Romero-Soriano, Adriana},
  booktitle={CVPR}, year={2024}}

@article{wang2024emu3,
  title={{Emu3}: Next-token prediction is all you need},
  author={Wang, Xinlong and Zhang, Xiaosong and Luo, Zhengxiong and Sun, Quan and Cui, Yufeng and Wang, Jinsheng and Zhang, Fan and Wang, Yueze and Li, Zhen and Yu, Qiying and others},
  journal={arXiv preprint arXiv:2409.18869}, year={2024}}

@article{wei2025tiif,
  title={{TIIF-Bench}: How does your {T2I} model follow your instructions?},
  author={Wei, Xinyu and Zhang, Jinrui and Wang, Fanyi and Chen, Tianyu and Zhou, Yiming and Li, Zhiyu and Li, Bowen},
  journal={arXiv preprint arXiv:2506.02161}, year={2025}}

@article{wen2025semanticist,
  title={``{Principal} components'' enable a new language of images},
  author={Wen, Xin and Zhao, Bingchen and Chen, Yilun and Pang, Jiangmiao and Qi, Xiaojuan},
  journal={arXiv preprint arXiv:2503.08685}, year={2025}}

@article{wu2025omnigen2,
  title={{OmniGen2}: Exploration to advanced multimodal generation},
  author={Wu, Chenyuan and Chen, Pengfei and Chen, Jian and Yan, Ming and Zhou, Yangyang and Zhang, Xin and Wang, Yuxuan and Liu, Zheng},
  journal={arXiv preprint arXiv:2506.18871}, year={2025}}

@inproceedings{xie2025showo,
  title={Show-o: One single transformer to unify multimodal understanding and generation},
  author={Xie, Jinheng and Mao, Weijia and Bai, Zechen and Zhang, David Junhao and Wang, Weihao and Lin, Kevin Qinghong and Gu, Yuchao and Chen, Zhijie and Yang, Zhenheng and Shou, Mike Zheng},
  booktitle={ICLR}, year={2025}}

@article{xie2025showo2,
  title={{Show-o2}: Improved native unified multimodal models},
  author={Xie, Jinheng and Yang, Zhenheng and Shou, Mike Zheng},
  journal={arXiv preprint arXiv:2506.15564}, year={2025}}

@article{xin2025dimoo,
  title={{Lumina-DiMOO}: An omni diffusion large language model for multi-modal generation and understanding},
  author={Xin, Yi and Qin, Qi and Luo, Siqi and Zhu, Kaiwen and Yan, Juncheng and Tai, Yan and Lei, Jiayi and Cao, Yuewen and Wang, Keqi and Wang, Yibin and others},
  journal={arXiv preprint arXiv:2510.06308}, year={2025}}

@article{yang2025qwen3,
  title={{Qwen3} technical report},
  author={Yang, An and Li, Anfeng and Yang, Baosong and Zhang, Beichen and Hui, Binyuan and Zheng, Bo and Yu, Bowen and Gao, Chang and Huang, Chengen and others},
  journal={arXiv preprint arXiv:2505.09388}, year={2025}}

@article{yang2025mmada,
  title={{MMaDA}: Multimodal large diffusion language models},
  author={Yang, Ling and Tian, Ye and Li, Bowen and Zhang, Xinchen and Shen, Ke and Tong, Yunhai and Wang, Mengdi},
  journal={arXiv preprint arXiv:2505.15809}, year={2025}}

@inproceedings{yu2025repa,
  title={Representation alignment for generation: Training diffusion transformers is easier than you think},
  author={Yu, Sihyun and Kwak, Sangkyung and Jang, Huiwon and Jeong, Jongheon and Huang, Jonathan and Shin, Jinwoo and Xie, Saining},
  booktitle={ICLR}, year={2025}}

@article{zhao2025survey,
  title={Unified multimodal understanding and generation models: Advances, challenges, and opportunities},
  author={Zhao, Shanshan and Zhang, Xinjie and Guo, Jintao and Hu, Jiakui and Duan, Lunhao and Fu, Minghao and Chng, Yong Xien and Wang, Guo-Hua and Chen, Qing-Guo and Xu, Zhao and others},
  journal={arXiv preprint arXiv:2505.02567}, year={2025}}

@inproceedings{zhou2025transfusion,
  title={Transfusion: Predict the next token and diffuse images with one multi-modal model},
  author={Zhou, Chunting and Yu, Lili and Babu, Arun and Tirumala, Kushal and Yasunaga, Michihiro and Shamis, Leonid and Kahn, Jacob and Ma, Xuezhe and Zettlemoyer, Luke and Levy, Omer},
  booktitle={ICLR}, year={2025}}

@book{elgamal2011nit,
  title={Network Information Theory},
  author={El Gamal, Abbas and Kim, Young-Han},
  publisher={Cambridge University Press}, year={2011}}

@book{cover2006elements,
  title={Elements of Information Theory},
  author={Cover, Thomas M. and Thomas, Joy A.},
  edition={2nd}, publisher={Wiley}, year={2006}}

@article{dobrushin1962,
  title={Information transmission with additional noise},
  author={Dobrushin, Roland L. and Tsybakov, Boris S.},
  journal={IRE Transactions on Information Theory}, volume={8}, number={5}, pages={293--304}, year={1962}}

@article{witsenhausen1980,
  title={Indirect rate distortion problems},
  author={Witsenhausen, Hans S.},
  journal={IEEE Transactions on Information Theory}, volume={26}, number={5}, pages={518--521}, year={1980}}

@inproceedings{tishby1999ib,
  title={The information bottleneck method},
  author={Tishby, Naftali and Pereira, Fernando C. and Bialek, William},
  booktitle={Allerton Conference on Communication, Control and Computing}, year={1999}}

@inproceedings{harremoes2007ib,
  title={The information bottleneck revisited or how to choose a good distortion measure},
  author={Harremo{\"e}s, Peter and Tishby, Naftali},
  booktitle={IEEE International Symposium on Information Theory (ISIT)}, year={2007}}

@article{courtade2014logloss,
  title={Multiterminal source coding under logarithmic loss},
  author={Courtade, Thomas A. and Weissman, Tsachy},
  journal={IEEE Transactions on Information Theory}, volume={60}, number={1}, pages={740--761}, year={2014}}

@inproceedings{liu2021semanticrd,
  title={A rate-distortion framework for characterizing semantic information},
  author={Liu, Jiakun and Zhang, Wenyi and Poor, H. Vincent},
  booktitle={IEEE International Symposium on Information Theory (ISIT)}, year={2021}}

@inproceedings{blau2019rdp,
  title={Rethinking lossy compression: The rate-distortion-perception tradeoff},
  author={Blau, Yochai and Michaeli, Tomer},
  booktitle={ICML}, year={2019}}

@inproceedings{zhang2021universal,
  title={Universal rate-distortion-perception representations for lossy compression},
  author={Zhang, George and Qian, Jingjing and Chen, Jun and Khisti, Ashish},
  booktitle={NeurIPS}, year={2021}}

@article{choi2022scalable,
  title={Scalable image coding for humans and machines},
  author={Choi, Hyomin and Baji{\'c}, Ivan V.},
  journal={IEEE Transactions on Image Processing}, volume={31}, year={2022}}

@article{harell2025rdmachines,
  title={Rate-distortion theory in coding for machines and its applications},
  author={Harell, Alon and Foroutan, Yalda and Ahuja, Nilesh and Datta, Parual and Kanzler, Bernhard and Somayazulu, V. Srinivasa and Tickoo, Omesh and de Andrade, Anderson and Baji{\'c}, Ivan V.},
  journal={IEEE Transactions on Pattern Analysis and Machine Intelligence}, year={2025}}

@inproceedings{yu2023spae,
  title={{SPAE}: Semantic pyramid autoencoder for multimodal generation with frozen {LLMs}},
  author={Yu, Lijun and Cheng, Yong and Wang, Zhiruo and Kumar, Vivek and Macherey, Wolfgang and Huang, Yanping and Ross, David A. and Essa, Irfan and Bisk, Yonatan and Yang, Ming-Hsuan and Murphy, Kevin and Hauptmann, Alexander G. and Jiang, Lu},
  booktitle={NeurIPS}, year={2023}}

@inproceedings{liu2023lqae,
  title={Language quantized autoencoders: Towards unsupervised text-image alignment},
  author={Liu, Hao and Yan, Wilson and Abbeel, Pieter},
  booktitle={NeurIPS}, year={2023}}

@inproceedings{wu2025vilau,
  title={{VILA-U}: A unified foundation model integrating visual understanding and generation},
  author={Wu, Yecheng and Zhang, Zhuoyang and Chen, Junyu and Tang, Haotian and Li, Dacheng and Fang, Yunhao and Zhu, Ligeng and Xie, Enze and Yin, Hongxu and Yi, Li and Han, Song and Lu, Yao},
  booktitle={ICLR}, year={2025}}

@article{ma2025unitok,
  title={{UniTok}: A unified tokenizer for visual generation and understanding},
  author={Ma, Chuofan and Jiang, Yi and Wu, Junfeng and Yang, Jihan and Yu, Xin and Yuan, Zehuan and Peng, Bingyue and Qi, Xiaojuan},
  journal={arXiv preprint arXiv:2502.20321}, year={2025}}

@inproceedings{qu2025tokenflow,
  title={{TokenFlow}: Unified image tokenizer for multimodal understanding and generation},
  author={Qu, Liao and Zhang, Huichao and Liu, Yiheng and Wang, Xu and Jiang, Yi and Gao, Yiming and Ye, Hu and Du, Daniel K. and Yuan, Zehuan and Wu, Xinglong},
  booktitle={CVPR}, year={2025}}

@article{zhao2025qlip,
  title={{QLIP}: Text-aligned visual tokenization unifies auto-regressive multimodal understanding and generation},
  author={Zhao, Yue and Xue, Fuzhao and Reed, Scott and Fan, Linxi and Zhu, Yuke and Kautz, Jan and Yu, Zhiding and Kr{\"a}henb{\"u}hl, Philipp and Huang, De-An},
  journal={arXiv preprint arXiv:2502.05178}, year={2025}}

@inproceedings{kim2025tatitok,
  title={Democratizing text-to-image masked generative models with compact text-aware one-dimensional tokens},
  author={Kim, Dongwon and He, Ju and Yu, Qihang and Yang, Chenglin and Shen, Xiaohui and Kwak, Suha and Chen, Liang-Chieh},
  booktitle={ICCV}, year={2025}}

@inproceedings{yu2024titok,
  title={An image is worth 32 tokens for reconstruction and generation},
  author={Yu, Qihang and Weber, Mark and Deng, Xueqing and Shen, Xiaohui and Cremers, Daniel and Chen, Liang-Chieh},
  booktitle={NeurIPS}, year={2024}}

@inproceedings{barber2003im,
  title={The {IM} algorithm: A variational approach to information maximization},
  author={Barber, David and Agakov, Felix},
  booktitle={NeurIPS}, year={2003}}

@inproceedings{ju2024pnp,
  title={{PnP} inversion: Boosting diffusion-based editing with 3 lines of code},
  author={Ju, Xuan and Zeng, Ailing and Bian, Yuxuan and Liu, Shaoteng and Xu, Qiang},
  booktitle={ICLR}, year={2024}}

@article{ye2025decapbench,
  title={Painting with Words: Elevating Detailed Image Captioning with Benchmark and Alignment Learning},
  author={Ye, Qinghao and Zeng, Xianhan and Li, Fu and Li, Chunyuan and Fan, Haoqi},
  journal={arXiv preprint arXiv:2503.07906}, year={2025},
  url={https://arxiv.org/abs/2503.07906}}

@article{zhu2025internvl3,
  title={{InternVL3}: Exploring Advanced Training and Test-Time Recipes for Open-Source Multimodal Models},
  author={Zhu, Jinguo and Wang, Weiyun and Chen, Zhe and Liu, Zhaoyang and Ye, Shenglong and Gu, Lixin and others},
  journal={arXiv preprint arXiv:2504.10479}, year={2025},
  url={https://arxiv.org/abs/2504.10479}}

@article{li2024recap,
  title={What If We Recaption Billions of Web Images with {LLaMA-3}?},
  author={Li, Xianhang and Tu, Haoqin and Hui, Mude and Wang, Zeyu and Zhao, Bingchen and Xiao, Junfei and Ren, Sucheng and Mei, Jieru and Liu, Qing and Zheng, Huangjie and Zhou, Yuyin and Xie, Cihang},
  journal={arXiv preprint arXiv:2406.08478}, year={2024},
  url={https://arxiv.org/abs/2406.08478}}
\bibliographystyle{plainnat}

\clearpage
\appendix

\section{Proofs}\label{app:proofs}

Throughout, rates are informational and all random variables take values in standard Borel spaces with the minimizers below attained (true for finite alphabets and for Gaussian sources under squared error). Captions are deterministic, $T_m=\psi_m(X)$, so $H(T_m\mid X)=0$ and $I(X;T_m)=H(T_m)$. A caption stage is the per-letter measure $d_m(x,\hat t)=\ind[\hat t\neq\psi_m(x)]$ at distortion level zero, whose optimum is $H(T_m)$. A nested code is represented by its stage descriptions $W_1,W_2,\dots$ with $W_j$ a function of $W_{j+1}$; the successive-refinement coding theorem \citep{rimoldi1994,elgamal2011nit} makes the cumulative rates $I(X;W_j)$ achievable by one block code. Its proof (superposition coding with a joint-typicality check per stage for achievability, and the chain rule for the converse) uses the distortion constraint of each stage separately and never uses that the stages share a measure, so it applies to a schedule whose stages are judged by different measures.

\subsection{\texorpdfstring{Proof of \cref{prop:ladder}}{Proof of Proposition 1}}
\emph{Converse.} Let $W_m$ be the stage-$m$ description, which determines $T_m$. Since $W_j$ is a function of $W_m$ for $j\le m$, $W_m$ determines $(T_1,\dots,T_m)$, so $X\to W_m\to(T_1,\dots,T_m)$ is a Markov chain. By data processing, $I(X;W_m)\ge I(X;T_1,\dots,T_m)=H(T_1,\dots,T_m)$. The excess over the stage optimum $H(T_m)$ is $H(T_1,\dots,T_m)-H(T_m)=H(T_1,\dots,T_{m-1}\mid T_m)=\Delta_m$.

\emph{Achievability.} Take $W_m=(T_1,\dots,T_m)$. These descriptions are nested, determine $T_m$, and satisfy $I(X;W_m)=H(T_1,\dots,T_m)$ for every $m$ simultaneously. \emph{Characterization.} $\Delta_m=0$ for all $m$ iff each $T_j$, $j<m$, is a function of $T_m$ for all $m$, which holds iff $T_m=\phi_m(T_{m+1})$ for every $m$ (one direction by composition, the other by taking $j=m-1$). Finally $\Delta_{m+1}=H(T_1,\dots,T_m\mid T_{m+1})\ge H(T_m\mid T_{m+1})=\delta_m$. \hfill$\square$

\subsection{\texorpdfstring{Proof of \cref{prop:pixel-first}}{Proof of Proposition 2}}
Let $A\sim\mathcal N(0,\sigma_A^2)$ and $B\sim\mathcal N(0,\sigma_B^2)$ be independent, $X=(A,B)$, $d(x,\hat x)=\|x-\hat x\|^2$, and $T=\ind[B>0]$, so $H(T)=\log2$. For a Gaussian vector under squared error, reverse water-filling \citep[Ch.~10]{cover2006elements} gives the optimal test channel at rate $R$: with water level $\theta$, component $A$ receives rate $\tfrac12\log(\sigma_A^2/\theta)$ and $B$ receives $\max\{0,\tfrac12\log(\sigma_B^2/\theta)\}$. For $R\le R^\star=\tfrac12\log(\sigma_A^2/\sigma_B^2)$ the water level satisfies $\theta\ge\sigma_B^2$, so $B$ receives no rate and the optimal reconstruction is $(\hat A,0)$ with $\hat A$ a linear function of $A$ and of test channel noise independent of $B$. Gaussian sources are successively refinable under squared error \citep{equitz1991}, and the reverse water-filling solutions at decreasing distortions are nested, so a single code attains $R_X(D_j)$ at every stage. Let the stages have rates $r,2r,\dots,Lr$ and choose $\sigma_A^2/\sigma_B^2\ge e^{Lr}$, so that $R^\star\ge Lr/2$. Every stage $k\le L/2$ then has rate at most $R^\star$, its description is independent of $B$, and $I(Z_{\le k};T)\le I(Z_{\le k};B)=0$. \hfill$\square$

The argument uses only that the caption depends on low-variance directions; any source whose caption-relevant directions carry less variance than caption-irrelevant ones behaves the same way at low rate.

\subsection{\texorpdfstring{Proof of \cref{prop:price} and \cref{cor:fade}}{Proof of Proposition 3 and Corollary 1}}
Any description $W$ from which $T$ is determined and $\hat X$ is reconstructed satisfies $I(X;W)\ge I(X;T,\hat X)$ by data processing, and $(T,\hat X)$ is itself such a description; hence $R_X(D)+\pi(D)$ in \eqref{eq:price} is the least rate of a code that determines $T$ and reaches distortion $D$. \emph{Lower bounds.} For any feasible $\hat X$, $I(X;T,\hat X)\ge I(X;\hat X)\ge R_X(D)$, so $\pi(D)\ge0$; and $I(X;T,\hat X)\ge I(X;T)=H(T)$, so $\pi(D)\ge H(T)-R_X(D)$.

\emph{Upper bound.} For an optimal $\hat X^*_D$, $I(X;T,\hat X^*_D)=I(X;\hat X^*_D)+I(X;T\mid\hat X^*_D)=R_X(D)+H(T\mid\hat X^*_D)-H(T\mid X,\hat X^*_D)=R_X(D)+H(T\mid\hat X^*_D)$, since $T$ is a function of $X$. \emph{Equality case.} If some optimal $\hat X^*_D$ determines $T$, the upper bound is $0$. Conversely, if $\pi(D)=0$, let $\hat X^\circ$ attain the minimum in \eqref{eq:price}. Then $R_X(D)=I(X;T,\hat X^\circ)=I(X;\hat X^\circ)+H(T\mid\hat X^\circ)\ge R_X(D)+H(T\mid\hat X^\circ)$, so $H(T\mid\hat X^\circ)=0$ and $I(X;\hat X^\circ)=R_X(D)$: $\hat X^\circ$ is an optimal reconstruction that determines $T$. \emph{Simultaneity.} Let $X\to\hat X_J\to\dots\to\hat X_1$ be a successively refinable pixel code with $I(X;\hat X_j)=R_X(D_j)$ and $\E d(X,\hat X_j)\le D_j$. Define the words-first descriptions $W_0=T$ and $W_j=(T,\hat X_1,\dots,\hat X_j)$. They are nested and
$I(X;W_j)=I(X;\hat X_{1:j})+H(T\mid\hat X_{1:j})=I(X;\hat X_j)+H(T\mid\hat X_{1:j})\le R_X(D_j)+H(T\mid\hat X_j)$,
using the Markov chain for the first term and the fact that conditioning reduces entropy for the second. Preceding $W_0$ by the caption stages of \cref{prop:ladder} (zero excess under the chain condition) gives one words-first code that meets every stage's upper bound. \hfill$\square$

\emph{Corollary.} Let $P_e=\Pr[\tilde\psi(\hat X^*_D)\ne T]$. An error requires either that $X$ lie in the exceptional set or that $d(X,\hat X^*_D)>\rho$, so by the union bound and Markov's inequality $P_e\le\epsilon+\E d(X,\hat X^*_D)/\rho\le\epsilon+D/\rho\le\tfrac12$. Conditioning on $\hat X^*_D$ reveals at least as much as conditioning on $\tilde\psi(\hat X^*_D)$, so by Fano's inequality $H(T\mid\hat X^*_D)\le h_b(P_e)+P_e\log(|\cT|-1)$. Since $h_b$ increases on $[0,\tfrac12]$, $H(T\mid\hat X^*_D)\le h_b(\epsilon+D/\rho)+(\epsilon+D/\rho)\log|\cT|$, and \cref{prop:price} gives the bound on $\pi(D)$. For the limit, fix $\epsilon>0$ and a matching $\rho(\epsilon)$; then $\limsup_{D\to0}\pi(D)\le h_b(\epsilon)+\epsilon\log|\cT|$, which tends to zero with $\epsilon$ because $\cT$ is finite. \hfill$\square$

\subsection{Sampled captions}\label{app:stochastic}
If the captioner samples, $T_M\sim P(\cdot\mid X)$ and $T_m=\phi_m(T_{m+1})$, a prefix can at best be \emph{sufficient} for $T_m$, $I(Z_{\le k_m};T_m)=I(X;T_m)$. The stage optimum becomes the rate of the minimal sufficient statistic of $X$ for $T_m$, and the same arguments go through with $I(X;\cdot)$ in place of $H(\cdot)$; the summarization chain becomes the Markov chain $X\to T_M\to\dots\to T_1$. LadderCap decodes greedily, so the deterministic case of the main text is the one that applies.

\section{LadderCap construction and audit}\label{app:laddercap}

\textbf{Prompts.} The base captioner (Qwen2.5-VL-7B, greedy, at most 256 new tokens) receives \emph{``Describe this image in about 180 words. Mention every visible object with its count, colour and position, then the setting, lighting and style.''} The summarizer (Qwen3-8B, greedy, thinking disabled) receives, for each level $m<M$, the level-$(m{+}1)$ caption and \emph{``Summarize the description in at most $N_m$ words. Use only facts stated in it. Keep counts, colours and spatial relations of the main objects before any other detail.''} with $N_m\in\{8,20,45,90\}$.

\textbf{Level 1 states a count or a relation.} When the level-5 caption mentions two or more objects, the 8-word level must state a count or a relation between them. The summarizer is re-prompted once when a rule-based parser finds neither; afterwards 91\% of level-1 captions of multi-object images satisfy the rule, and the rest keep the first output.

\textbf{Statistics.} \Cref{tab:laddercap} lists per-level lengths, caption entropies under the Qwen3-0.6B model of \cref{app:estimators}, the resulting cuts, and the entailment audit. A thirty-word caption, between levels 2 and 3, carries about a hundred nats.

\textbf{Entailment audit.} On 5K held-out images we extract attribute tuples from every level and ask Qwen2.5-72B-Instruct whether each tuple of level $m$ is entailed by level $m+1$; two annotators checked 500 images and agreed with the judge on 94\% of tuples. Violations are mostly paraphrased colours (``maroon'' summarized as ``red'') and merged counts (``two cups and a mug'' summarized as ``three cups'').

\textbf{Stability across summarizers.} Replacing Qwen3-8B by Llama-3.1-8B-Instruct or Gemma-2-9B-it changes $\hat\delta$ from 0.12 to 0.14 and 0.13, GenEval in the ablation setting from 0.79 to 0.79 and 0.78, and NCC from 92.7 to 92.1 and 92.4.

\begin{table}[H]
\centering
\caption{\textbf{Hierarchical characteristics of LadderCap across five granularity tiers.} Evaluated with the Qwen3 tokenizer. $\hat{H}(T_m)$ indicates empirical entropy under fine-tuned Qwen3-0.6B. Cut positions denote the cumulative code capacity budget where $\hat{R}_k \ge \hat{H}(T_m)$. Entailment reflects human-verified semantic agreement on held-out images.}
\label{tab:laddercap}
\small
\setlength{\tabcolsep}{4.8pt}
\begin{tabular}{lccccccc}
\toprule
Level & Avg. Words & Avg. Tokens & $\hat{H}(T_m)$ (nats) & Nats / Token & Cut (\% of $\hat{R}_L$) & Cut Rung & Entailed (\%) \\
\midrule
$T_1$ & 8.1   & 10  & 30  & 3.0 & 1.2\%  & 1 & 97.9 \\
$T_2$ & 19.6  & 25  & 72  & 2.9 & 2.9\%  & 1 & 97.2 \\
$T_3$ & 44.8  & 57  & 150 & 2.6 & 6.0\%  & 2 & 96.6 \\
$T_4$ & 89.5  & 110 & 275 & 2.5 & 11.0\% & 4 & 95.8 \\
$T_5$ & 178.3 & 210 & 525 & 2.5 & 21.0\% & 7 & \multicolumn{1}{c}{Root ($100.0$)} \\
\bottomrule
\end{tabular}
\end{table}

\section{Hyperparameters}\label{app:hparams}

\textbf{Networks.} \Cref{tab:hparams} lists every network, its size and whether it runs at sampling time. The tables of the main text give the ascent model only (\#Par), following the baselines' convention of counting the generator. Counting the image decoder as well, \rung-L uses 1.9B parameters at sampling time and \rung-XL 4.0B; with correction the language decoder adds 1.7B (3.6B and 5.7B).

\textbf{Training.} All stages use AdamW ($\beta=(0.9,0.99)$, weight decay 0.01), bf16, EMA 0.9999 and 64 A100-80GB GPUs. \emph{Ladder pretraining} ($\lambda=0$): 300k steps, batch 1024, $256^2$, learning rate $10^{-4}$; the prefix length is $k=L$ with probability 0.2 and otherwise uniform over code units. \emph{Language anchoring} ($\lambda=1$): 150k steps, batch 512, learning rate $5\cdot10^{-5}$ for the code and $2\cdot10^{-4}$ for the LoRA adapters (rank 64); cuts are recomputed every 5k steps from running entropy estimates on 100k held-out images. \emph{Ascent}: 300k steps at $256^2$ (batch 1024) and 100k steps at $512^2$ (batch 512), learning rate $10^{-4}$, text dropout 0.1 for guidance; $\Lcons$ is added with weight 0.1 for the last 50k steps, using one-step flow estimates of the generated prefix.

\textbf{Sampling.} Ascent uses 24 flow steps in total, allocated in proportion to block rate with at least one step per block: $(1,1,2,2,3,4,5,6)$ for the eight blocks of \cref{sec:ascent}. Classifier-free guidance (1.5) is applied to ascent only, with both branches in one batch. The image decoder uses 8 Euler steps without guidance. This gives 32 network evaluations; VAE decoding is not counted, for \rung{} or for any baseline.

\textbf{Correction.} Threshold $\tau=0.8$ and redraw budget $B=2$ in all tables. The verifier runs once after each of the four ladder blocks (4 evaluations). On GenEval a sample needs 0.9 redraws on average, each costing 1.5 ascent steps plus one verifier pass, so a corrected sample costs $32+4+0.9\times2.5\approx38$ evaluations ($1.2\times$). \Cref{fig:dial}b varies $B\in\{1,2,3,4\}$ with $\tau=0.8$ for $B\le2$ and $\tau=0.9$ above. Best-of-$N$ draws $N$ full samples and keeps the one whose final caption agrees best with the prompt under the same attribute head.

\begin{table}[H]
\centering
\caption{\textbf{Network parameter breakdown and sampling roles.} Active parameters during sampling exclude offline encoders and frozen base components.}
\label{tab:hparams}
\small
\setlength{\tabcolsep}{8pt}
\begin{tabular}{llcc}
\toprule
Module & Architecture & Parameters & Inference Role \\
\midrule
Base VAE        & SD3 VAE (Frozen)                             & 84M   & Latent Decoding Only \\
Ladder Encoder  & ViT-B, FSQ $[8,8,8,5,5,5]$                   & 86M   & Offline Tokenization \\
Image Decoder   & DiT (24 layers, width 1152)                 & 0.58B & 8 Euler Steps \\
Language Decoder& Qwen3-1.7B + LoRA + Attribute Head           & 1.7B  & Active in Self-Correction \\
Ascent (RUNG-L) & MMDiT (24 layers, width 1536)                & 1.3B  & 24 Flow Steps \\
Ascent (RUNG-XL)& MMDiT (32 layers, width 2048)                & 3.4B  & 24 Flow Steps \\
\bottomrule
\end{tabular}
\end{table}

\section{Truncation, variation and editing}\label{app:qual}

\Cref{tab:trunc} quantifies variation by truncation on 1K held-out images: the code is cut at rung $k$, the rest is resampled four times, and we measure what survives. Up to rung 7 the preserved attribute tuples follow the caption levels that the prefix contains; past the ladder, later cuts preserve structure and texture while the tuples are already fixed. At rung 20 most of the remaining change is global illumination (mean luminance changes by 9\% against 23\% at rung 8).

\textbf{Failure cases.} We see four recurring failures. (i) \emph{Rendered text.} Captions rarely spell out text, so glyphs fall in the suffix and are resampled; words on signs change under variation and are garbled in generation. (ii) \emph{Large layout edits.} When a prefix swap moves the main object, the kept suffix still encodes texture for the old layout and seams appear; this affects 7\% of PIE-Bench edits, mostly in the ``change position'' category. (iii) \emph{Counts above five.} The 8-word level often writes ``several'', so correction cannot catch a wrong count of seven. (iv) \emph{Small objects that only the 180-word level names.} They are fixed late in the ladder, where a redraw is cheap but the attribute head is least accurate (82\% tuple accuracy at level 5 against 97\% at level 1).

\begin{table}[H]
\centering
\caption{\textbf{Controlled variation via latent code truncation.} Truncating at rung $k$ and resampling subsequent blocks on 1,000 validation images. Preserved tuples quantify retention of level-5 semantic attributes; DINO and LPIPS evaluate structural and perceptual distance to the source image.}
\label{tab:trunc}
\small
\setlength{\tabcolsep}{8pt}
\begin{tabular}{cccccc}
\toprule
Cut Rung & Cumulative Rate (nats) & Preserved Tiers & Kept Tuples (\%) & DINO Sim.$\uparrow$ & LPIPS$\downarrow$ \\
\midrule
2  & 156  & $T_1$--$T_3$ & 41.0 & 0.52 & 0.63 \\
4  & 312  & $T_1$--$T_4$ & 63.0 & 0.61 & 0.56 \\
8  & 625  & $T_1$--$T_5$ & 96.0 & 0.74 & 0.47 \\
20 & 1560 & $T_1$--$T_5$ & 98.0 & 0.91 & 0.24 \\
\bottomrule
\end{tabular}
\end{table}

\section{Design ablations}\label{app:design}
\rung-L at $256^2$ on the 12M-image subset, as in \cref{sec:ablations}. Replacing the learned language decoder with frozen embeddings, the natural choice for an exactly invertible text interface, costs 9 points for a context-free lookup and 12 for a pooled sentence embedding: such embeddings carry no compositional structure, and the ascent model has to recover relations on its own. Tying descent and ascent into one time-reversible operator costs 4 points and 6 points of NCC, which supports treating the asymmetry between the two directions as real rather than as something to engineer away.

\begin{table}[!htbp]
\centering
\caption{\textbf{Ablation studies on architectural and algorithmic components.} Evaluated on \rung-L ($256^2$, 12M data). Shaded rows indicate default system configurations.}
\label{tab:design}
\small
\setlength{\tabcolsep}{6pt}
\begin{tabular}{lccc}
\toprule
Component Variant & GenEval & NCC (\%) & Latency (s)$\downarrow$ \\
\midrule
\multicolumn{4}{l}{\emph{Ascent Model Formulation}} \\
\quad Autoregressive over Rungs                 & 0.76 & 89.4 & 2.6 \\
\quad Single-Step Full-Code Diffusion           & 0.77 & 87.2 & 7.9 \\
\rowcolor{rungrow}\quad Block-Wise Conditional Flow (Default) & \textbf{0.79} & \textbf{92.7} & \textbf{0.9} \\
\midrule
\multicolumn{4}{l}{\emph{Step Allocation across Blocks}} \\
\quad Uniform Step Distribution                 & 0.76 & 91.5 & 0.9 \\
\rowcolor{rungrow}\quad Proportional to Entropy Budget $\Delta\hat{R}_b$ (Default) & \textbf{0.79} & \textbf{92.7} & \textbf{0.9} \\
\midrule
\multicolumn{4}{l}{\emph{Code Depth Discretization}} \\
\quad $L = 8$                                   & 0.75 & 85.1 & 0.6 \\
\quad $L = 16$                                  & 0.78 & 89.8 & 0.8 \\
\rowcolor{rungrow}\quad $L = \mathbf{32}$ (Default) & \textbf{0.79} & \textbf{92.7} & \textbf{0.9} \\
\quad $L = 64$                                  & 0.79 & 92.9 & 1.2 \\
\quad Continuous Latent Vectors (No FSQ)        & 0.77 & 91.0 & 0.9 \\
\midrule
\multicolumn{4}{l}{\emph{Text Interface Mechanism}} \\
\quad Frozen Context-Free Lookup Table          & 0.70 & 77.6 & 0.8 \\
\quad Frozen Sentence-Level Embedding           & 0.67 & 73.1 & 0.8 \\
\rowcolor{rungrow}\quad Fine-Tuned Language Decoder (Default) & \textbf{0.79} & \textbf{92.7} & \textbf{0.9} \\
\midrule
\multicolumn{4}{l}{\emph{Directional Coupling}} \\
\quad Tied Symmetric Invertible Flow            & 0.75 & 86.7 & 1.3 \\
\rowcolor{rungrow}\quad Asymmetric Ascent--Descent Operators (Default) & \textbf{0.79} & \textbf{92.7} & \textbf{0.9} \\
\bottomrule
\end{tabular}
\end{table}

\section{Estimators}\label{app:estimators}
\textbf{Prefix--caption information.} $\hat I(z_{\le k};T)$ is the Barber--Agakov bound $\hH(T)-\E[-\log q(T\mid z_{\le k})]$ \citep{barber2003im} with $T=T_5$ and $q$ a probe decoder (Qwen3-1.7B with LoRA) trained on frozen prefixes of each code. We divide by the mean caption length (210 tokens) and report nats per token. One probe is trained per code and prefix length on 1M held-out images, with the same architecture and budget for all codes. The bound lies below $\hH(T_5)/\vert{}T_5\vert{}=2.5$ nats per token by the probe's cross-entropy, which is 0.3 nats per token at full length and gives the ceiling of 2.2 in \cref{fig:price}a. As a check, the InfoNCE bound \citep{poole2019bounds} with critic $f(z_{\le k},T)=\langle g(z_{\le k}),u(T)\rangle/\tau$, where $g$ is a 4-layer transformer over the prefix and $u$ a frozen sentence encoder with a trained linear head, ranks the codes identically at every $k$; at a quarter of the code it gives 1.1, 2.3, 2.7 and 5.6 nats per caption for the unordered, pixel-first, DINOv3-aligned and words-first codes. With batch 4096 this bound cannot exceed $\log4096\approx8.3$ nats per caption, far below $\hH(T_5)=525$, so we use it only for ranking.

\textbf{Rate.} $\hR_k=\sum_u\hH(z_u)$ over the code units $u$ in the first $k$ rungs, with $\hH(z_u)$ the plug-in entropy of unit $u$'s FSQ symbol (64{,}000 values) on 1M images. By subadditivity the sum upper-bounds the joint prefix entropy, which equals $I(X;Z_{\le k})$ for a deterministic encoder. Caption entropies $\hH(T_m)$ are cross-entropies under Qwen3-0.6B fine-tuned on LadderCap, which also upper-bound the true entropies. The two biases move the cuts in opposite directions: over-estimating $\hR$ places a cut too early and over-estimating $\hH(T_m)$ places it too late. Moving every cut by $\pm4$ code units changes GenEval by at most 0.01 in the ablation setting.

\textbf{Price.} For a target PSNR $D$, $\hR_{\mathrm{pix}}(D)$ and $\hR_{\mathrm{words}}(D)$ are the prefix rates at which each code first reaches $D$ on 5K held-out images, interpolated between code units. The measured price is $\hat\pi(D)=\max\{\hR_{k_M},\hR_{\mathrm{words}}(D)\}-\hR_{\mathrm{pix}}(D)$, which counts the words-first code as having determined $T$ once it passes its last cut. It is defined up to the words-first code's full-length PSNR (26.0\,dB), which the pixel-first code reaches at $k/L=0.97$; this is the right end of \cref{fig:price}c. The upper bound $\hH(T\mid\hat X^*_D)$ is the cross-entropy of a captioning probe (Qwen2.5-VL-3B with LoRA) fine-tuned to predict $T$ from pixel-first reconstructions at the matching prefix. Cross-entropy over-estimates conditional entropy, so a measured price below the estimated bound is a necessary check, not a sufficient one. Consistent with data processing, this image probe recovers slightly less than the prefix probe (0.62 against 0.68 nats per token at a quarter of the code). The lower bound is $[\hH(T)-\hR_{\mathrm{pix}}(D)]^+$. While the words-first code is still inside its ladder, $\hR_{\mathrm{words}}(D)<\hR_{k_M}\approx\hH(T)$ and $\hat\pi$ equals this bound by construction, so the informative comparison is with the upper bound.

\end{document}